\documentclass{article}

\PassOptionsToPackage{numbers}{natbib}

\usepackage[preprint]{neurips_2019}

\makeatletter
\newcommand{\printfnsymbol}[1]{%
  \textsuperscript{\@fnsymbol{#1}}%
}
\makeatother

\PassOptionsToPackage{margin=1in}{geometry}

\usepackage[utf8]{inputenc} 
\usepackage[T1]{fontenc}    
\usepackage{hyperref}       
\usepackage{url}            
\usepackage{booktabs}       
\usepackage{amsfonts}       
\usepackage{nicefrac}       
\usepackage{microtype}      

\usepackage[margin=1in]{geometry}
\usepackage{amssymb}
\usepackage{amsmath}
\usepackage{algorithmic}
\usepackage{tabularx}
\usepackage{amsfonts}
\usepackage{algorithm2e}
\usepackage{tikz}
\usepackage{comment}
\usepackage{adjustbox}
\usepackage{subfigure}
\usepackage{sidecap}   
\usetikzlibrary{automata,positioning}

\newcommand{\states}{\mathcal{S}}
\newcommand{\actions}{\mathcal{A}}
\newcommand{\E}{\mathop{\mathbf{E}}}
\newcommand{\Var}{\mathrm{Var}}
\newcommand{\MC}{\mathrm{MC}}

\newcommand{\TDz}{\mathrm{TD}(0)}
\newcommand{\MSVE}{\mathrm{MSVE}}

\title{Adaptive Temporal-Difference Learning for Policy \\ Evaluation with Per-State Uncertainty Estimates}

\author{ {\bf Hugo Penedones} \thanks{These two authors contributed equally. Correspondence to \{hugopen, rikel\}@google.com} \\
DeepMind \\
\And
{\bf Carlos Riquelme} \printfnsymbol{1} \\
Google Research\\
\And
{\bf Damien Vincent}   \\
Google Research\\
\And
{\bf Hartmut Maennel}   \\
Google Research\\
\AND
{\bf Timothy Mann}   \\
DeepMind \\
\And
{\bf Andr\'e Barreto}   \\
DeepMind \\
\And
{\bf Sylvain Gelly}   \\
Google Research\\ 
\And
{\bf Gergely Neu}   \\
Universitat Pompeu Fabra \\
}

\begin{document}

\maketitle

\begin{abstract}
We consider the core reinforcement-learning problem of on-policy value function approximation from a batch of trajectory data, 
and focus on various issues of Temporal Difference (TD) learning and Monte Carlo (MC) policy evaluation. The two methods are known to achieve complementary bias-variance trade-off properties, with TD tending to achieve lower variance but potentially higher bias. In this paper, we argue that the larger bias of TD can be a result of the amplification of local approximation errors. We address this by proposing an algorithm that adaptively switches between TD and MC in each state, thus mitigating the propagation of errors. Our method is based on learned confidence intervals that detect biases of TD estimates. We demonstrate in a variety of policy evaluation tasks that this simple adaptive algorithm performs competitively with the best approach in hindsight, suggesting that learned confidence intervals are a powerful technique for adapting policy evaluation to use TD or MC returns in a data-driven way. 
\end{abstract}
\section{Introduction}

In reinforcement learning (RL) an agent must learn how to behave while interacting with an environment.
This challenging problem is usually formalized as the search for a decision policy---{\sl i.e.}, a mapping from states to actions---that maximizes the amount of reward received in the long run~\citep{sutton2018introduction}. Clearly, in order to carry out such a search we must be able to assess the quality of a given policy. This process, known as \emph{policy evaluation}, is the focus of the current paper.

A common way to evaluate a policy is to resort to the concept of \emph{value function}. Simply put, the value function of a policy associates with each state the expected sum of rewards, possibly discounted over time, that an agent following the policy from that state onwards would obtain. Thus, in this context the policy evaluation problem comes down to computing a policy's value function.

Perhaps the simplest way to estimate the value of a policy in a given state is to use \emph{Monte Carlo} (MC) returns: the policy is executed multiple times from the state of interest and the resulting outcomes are averaged ~\cite{sutton1998introduction}. Despite their apparent naivety, MC estimates enjoy some nice properties and have been advocated as an effective solution to the policy evaluation problem~\cite{amiranashvili2018analyzing}. 
Another way to address the policy evaluation problem is to resort to temporal-difference (TD) learning~\citep{sutton1988learning}. TD is based on the insight that the value of a state can be recursively defined based on other states' values~\citep{bellman1957dynamic}. Roughly speaking, this means that, when estimating the value of a state, instead of using an entire trajectory one uses the immediate reward plus the value of the next state. This idea of updating an estimate from another estimate allows the agent to learn online and incrementally.

Both MC and TD have advantages and disadvantages. From a statistical point of view, the estimates provided by MC are unbiased but may have high variance, while TD estimates show the opposite properties~\cite{sutton2018introduction}. As a consequence, the relative performance of the two methods depends on the amount of data available: while TD tends to give better estimates in small data regimes, MC often performs better with a large amount of data.
Since the amount of data that leads to MC outperforming TD varies from problem to problem, it is difficult to make an informed decision on which method to use in advance. It is also unlikely that the best choice will be the same for all states, not only because the number of samples associated with each state may vary but also because the characteristics of the value function itself may change across the state space.

Ideally, we would have a method that adjusts the balance between bias and variance per state based on the progress of the learning process. In this paper we propose an algorithm that accomplishes that by dynamically choosing between TD and MC before each value-function update.  
\emph{Adaptive TD} is based on a simple idea: if we have confidence intervals associated with states' values, we can decide whether or not to apply TD updates by checking if the resulting targets fall within these intervals. If the targets are outside of the confidence intervals we assume that the bias in the TD update is too high, and just apply an MC update instead. 

Although this idea certainly allows for many possible instantiations, in this work we focus on simple design choices.
Our experimental results cover a wide range of scenarios, from toy problems to Atari games, and they highlight the \emph{robustness} of the method, whose performance is competitive with the best of both worlds in most cases.
We hope this work opens the door to further developments in policy evaluation with function approximation in complex environments.
\newcommand{\val}{v}
\newcommand{\wh}{\widehat}

\section{The Problem}
\label{sec:problem}
This section formally introduces the problem of policy evaluation in Markov decision processes, as well as the two most basic approaches for tackling this fundamental problem: Monte-Carlo and Temporal-Difference learning. After the main definitions, we discuss the key advantages and disadvantages of these methods, which will enable us to state the main goals of our work.

\subsection{Policy Evaluation in Reinforcement Learning}

Let $M = \langle \states, \actions, P, r, \gamma, \mu_0 \rangle$ denote a Markov decision process (MDP) where $\states$ is the set of states, $\actions$ is the set of actions, and $P$ is the transition function so that, for all $s, s' \in \states$ and $a \in \actions$, $P(s'|s, a)$ denotes the probability of transitioning to $s'$ from state $s$ after taking action $a$.
Also, $r : \states \times \actions \rightarrow \mathbb{R}$ maps each pair (state, action) to its expected reward, $\gamma \in (0, 1]$ is the discount factor, and $\mu_0$ is the probability distribution over initial states.

Let $\pi: \states \to \mathcal{D}(\actions)$ be a policy, where $\mathcal{D}(\cdot)$ is the set of distributions over its argument set.
Assume at each state $s$ we sample $a \sim \pi(s)$.
The value function of $M$ at $s \in \states$ under policy $\pi$ is defined by
\begin{align}
 \val^{\pi}(s) = \E \left[ \sum_{t=0}^\infty \gamma^t \ r(S_t, \pi(S_t)) \middle| S_0 = s \right], \label{eqn:value_function}
\end{align}
where $S_{t+1}\sim P(\cdot|S_t,A_t)$ and $A_t\sim\pi(s)$.
We will drop the dependence of $r$ on $a$ for simplicity. 

Our goal is to recover value function $\val^{\pi}$ from samples collected by running policy $\pi$ in the MDP. As $\pi$ is fixed, we will simply use the notation $\val = \val^\pi$ below.
We consider trajectories collected on $M$ by applying $\pi$: $\tau = \langle (s_0, a_0, r_0), (s_1, a_1, r_1), \dots \rangle$. Given a collection of $n$ such trajectories $D_n = \{ \tau_i \}_{i=1}^n$, a policy evaluation algorithm outputs a function $\wh{V} : \states \rightarrow \mathbb{R}$.
We are interested in designing algorithms that minimize the Mean Squared Value Error of $\wh{V}$ defined as
\begin{equation}\label{eq:msve_def}
    \textrm{MSVE}(\wh{V}) = \E_{S_0 \sim \mu_0} \left[ \left( \val(S_0) - \wh{V}(S_0) \right)^2 \right].
\end{equation}
More specifically, we will search for an appropriate value estimate $\wh{V}$ within a fixed hypothesis set of functions $\mathcal{H} = \{ h : \states \rightarrow \mathbb{R} \}$, attempting to find an element with error comparable to $\min_{h \in \mathcal{H}} \textrm{MSVE}(h)$.
We mainly consider the set $\mathcal{H}$ of neural networks with a fixed architecture.

The key challenge posed by the policy evaluation problem is that the regression target $\val(s)$ in \eqref{eq:msve_def} is not directly observable. The algorithms we consider deal with this challenge by computing an appropriate regression target $T(s)$ and, instead, attempt to minimize $(T(s) - \wh{V}(s))^2$ as a function of $\wh{V}$, usually via stochastic optimization.

\subsection{Monte Carlo for Policy Evaluation}
The Monte-Carlo approach is based on the intuitive observation that the infinite discounted sum of rewards realized by running the policy from a state $s$ is an unbiased estimator of $v(s)$. This suggests that a reasonably good regression target can be constructed for all $s_t\in\tau_i$ as
\begin{equation}\label{eq:mc_return}
T_{\MC}\left(s_t^{(i)} \right) := \sum_{k=0}^{n_i-t-1} \gamma^k \ r \left(s_{t+k}^{(i)} \right),
\end{equation}
where $n_i$ is the length of trajectory $\tau_i$.
Thus, one viable approach for policy evaluation is to compute the minimizer within $h \in \mathcal{H}$ of 
$
     \sum_{i=1}^n \sum_{t=1}^{n_i} (T_{\MC}(s_t^{(i)}) - h(s_t^{(i)}))^2
$.
Risking some minor inconsistency\footnote{\cite{sutton1998introduction,sutton2018introduction} exclusively refer to the tabular version of the above method as Monte Carlo; this method is a natural generalization to general value-function classes.}
with the literature, we refer to this method as Monte Carlo policy evaluation.

\subsection{Temporal-Difference for Policy Evaluation}

Temporal-Difference algorithms are based on the fact that the value function should satisfy Bellman equations:
$v(s) = r(s) + \gamma \sum_{s'} P(s'|s,a) \ v(s')$
for all $s$, which suggests that a good estimate of the value function should minimize the squared error between the two sides of the above equation. In our framework, this can be formulated as using the regression target 
\begin{equation}\label{eq:td0_target}
T_{\TDz}\left(s_t^{(i)} \right) := 
r \left(s_{t}^{(i)} \right) + \gamma \ \wh{V} \left( s_{t+1}^{(i)} \right)
\end{equation}
to replace $v(s)$ in the objective of Equation~\eqref{eq:msve_def}. The practice of using the estimate $\wh V$ as part of the target is commonly referred to as ``bootstrapping'' within the RL literature. Again with a slight abuse of common terminology\footnote{This algorithm would be more appropriately called ``least squares TD'' or LSTD, following \cite{bradtke1996linear}, with the understanding that our method considers general (rather than linear) value-function classes.}, we will refer to this algorithm as TD(0), or just TD.

\begin{figure}[t]
\adjustbox{valign=t}{
\begin{minipage}[t]{0.32 \linewidth}
  \begin{center}
    \includegraphics[width=1.0\linewidth]{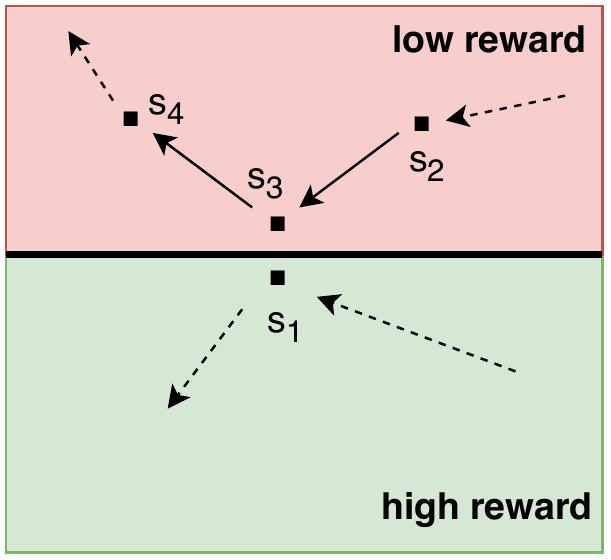}
  \end{center}
\end{minipage}}\hfill%
\adjustbox{valign=t}{
\begin{minipage}[t]{0.64 \linewidth}  
    \begin{tabularx}{1.0\linewidth}{XX}
    \includegraphics[width=1.0\linewidth]{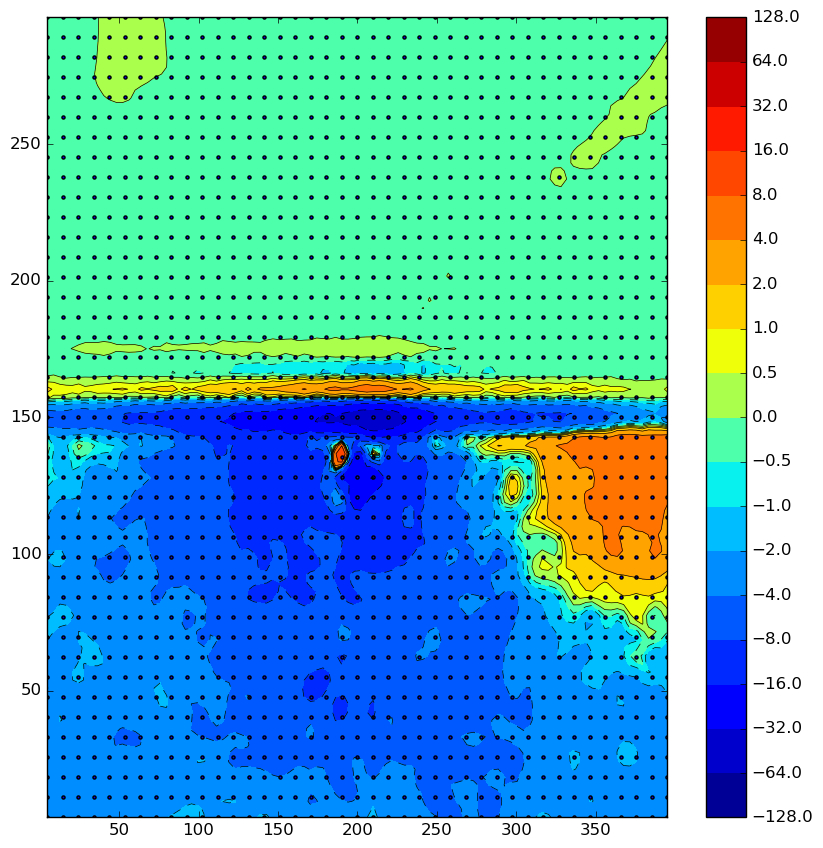} &
    \includegraphics[width=1.0\linewidth]{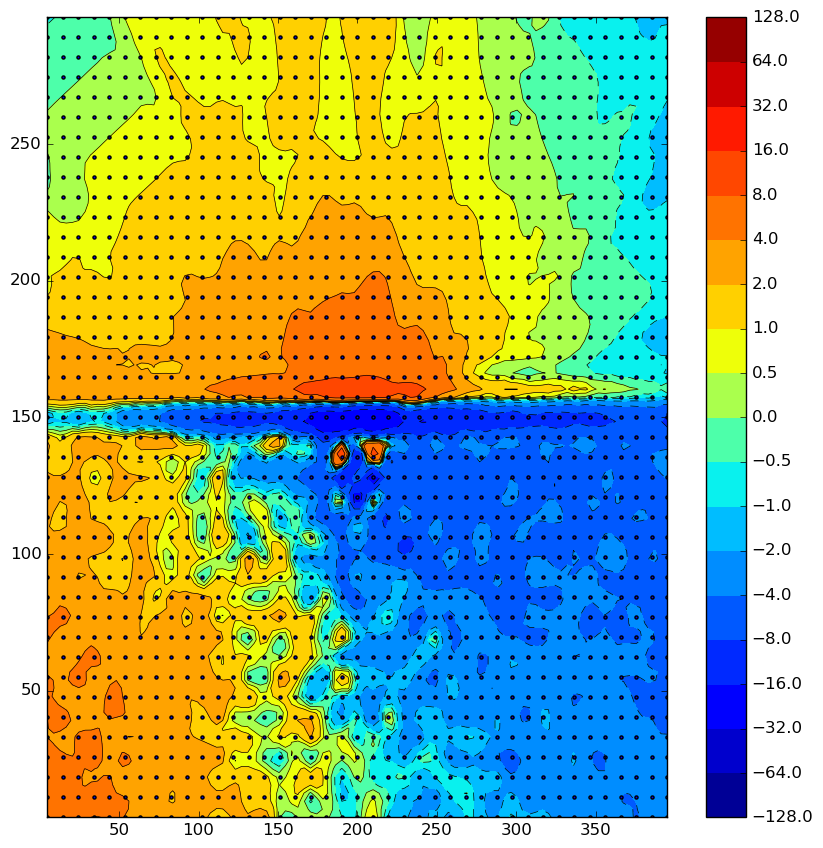}
    \end{tabularx}

\end{minipage}}%
\caption{
\textbf{Left.} Bootstrapping approximation errors.
\textbf{Center and Right.} MC (center) versus TD (right) on a simple environment with 2 rooms completely separated by a wall (see Map 2 in  Figure~\ref{fig:labyrinth_2d_value_functions_0_5}). 
  For each state $s$, the heatmaps show: $\hat{V}(s) - V(s)$. 
 The true value on the upper half of the plane is zero.
 MC overestimates the values of a narrow region right above the wall, due to function approximation limitations.
 With TD, these unavoidable approximation errors also occur, but things get worse when bootstrap updates propagate them to much larger regions (see right).
}
\label{fig:wall_examples}
\end{figure}

TD and Monte Carlo provide different target functions and, depending on the problem instance, each of them may offer some benefits.
In the tabular case, it is easy to see that Monte Carlo converges to the optimal solution with an infinite amount of data, since the targets concentrate around their true mean, the value function. However, the Monte Carlo targets can also suffer from large variance due to the excessive randomness of the cumulative rewards. On the other hand,  
TD can be shown to converge to the true value function in the same setting too \cite{sutton1988learning}, with the additional potential to converge faster due to the potential variance reduction in the updates. Indeed, when considering a fixed value function, the only randomness in the TD target is due to the immediate next state, whereas the MC target is impacted by the randomness of the entire trajectory.

Thus, the advantage of TD is more pronounced in low data regimes, or when the return distribution from a state has large variance.

The story may be different with function approximation: even in the limit of infinite data, both MC and TD are going to lead to biased estimates of the value function, due to the approximation error introduced by the function class $\mathcal{H}$. In the case of linear function classes, the biases of the two algorithms are well-understood; the errors in estimating the value functions can be upper-bounded in terms of the distance between the value function and the span of the features, with MC enjoying  tighter upper bounds than TD \citep{tsitsiklis1997analysis,scherrer2010should}. These results, however, do not provide a full characterization of the errors: even when considering linear function approximation, there are several known examples in the literature where TD provably outperforms MC and vice versa \citep{scherrer2010should}. Thus, the winner between the two algorithms will be generally determined by the particular problem instance we are tackling.

To see the intuitive difference between the behavior of the two algorithms, consider a situation where the true underlying value function $V$ has some sharp discontinuities that the class of functions at hand is not flexible enough to capture. In these cases, both methods suffer to fit the value function in some regions of the state space, even when we have lots of data. The errors, however, behave differently for the two methods: while the errors of Monte Carlo are localized to the regions with discontinuities due to directly fitting the data, TD bootstraps values from this problematic region, and thus propagates the errors even further. We refer to such errors arising due to discontinuities as \emph{leakage} \citep{penedones2018leakage}.

To illustrate the leakage phenomenon, consider the toy example in the left side of Figure~\ref{fig:wall_examples}.
There is a wall separating the green region (high reward), and the red region (low reward).
Assume that a function approximator $f$ will need to make a compromise to fit both $s_1$ and $s_3$ as they are close in the state space, even though no trajectory goes through both of them due to the wall.
For example, we can assume $f$ will over-estimate the true value of $s_3$.
Let us now examine a third state, $s_2$, that is in the red region, and such that there is a trajectory that goes from $s_2$ to $s_3$. 
The distance in state-space between $s_2$ and $s_3$ could be, in principle, arbitrarily large.
If we apply Monte Carlo updates, then the impact of the $s_1 \to s_3$ leakage on $s_2$ will be minimal.
Instead, the TD update for the value of $s_2$ will explicitly depend on the estimate for $s_3$, which is overestimated due to function approximation near a wall.
In the center and right side of Figure~\ref{fig:wall_examples}, we show the actual outcome of running TD and MC in such a setting.
TD dramatically propagates estimation errors far into the low-reward region as expected, while MC is way more robust and errors stay located very close to the wall.
Alternatively, in a smoother scenario, however, bootstrapping function approximation estimates can still be certainly beneficial and speed up learning.
We present and analyze a toy MDP in Section~\ref{sec:simple_example}.

The key observation of our work is that different \emph{regions of the state space} may be best suited for either TD or MC updates, amending the existing folk wisdom that TD and MC may ``globally'' outperform each other in different MDPs.
Accordingly, we set our goal as  designing a method that adaptively chooses a target depending on the properties of the value function around the specific state.

\section{The Adaptive-TD Algorithm}

In the previous sections, 
we saw that whether MC or TD is a better choice heavily depends on the specific scenario we are tackling, and on the family of functions we choose to fit the value function.
While in hindsight we may be able to declare a winner, in practice the algorithmic decision needs to be made at the beginning of the process.
Also, running both and picking the best-performing one in training time can be challenging,
as the compound variance of the return distribution over long trajectories may require extremely large validation sets to test the methods and choose, while this then limits the amount of available training data.
We aim to design a \emph{robust} algorithm which does not require any knowledge of the environment, and that dynamically adapts to both the geometry of the true value and transition functions, and to the function approximator it has at its disposal.  

In this section, we propose an algorithmic approach that aims to achieve the best of both the MC and TD worlds.
The core idea is to limit the bootstrapping power of TD to respect some hard limits imposed by an \emph{MC confidence interval}.
The main driver of the algorithm is TD, since it can significantly speed up learning in the absence of obstacles.
However, in regions of the state space where we somehow suspect that our estimates may be poor (e.g., near walls, close to big rewards, non-markovianity, partial observability, or after irrecoverable actions) we would like to be more conservative, and rely mostly on targets purely based on data. 
We explicitly control how conservative we would like to be by tuning the \emph{confidence level} $\alpha$ of the intervals: by letting $\alpha$ go to 1, the intervals become trivially wide, and we recover TD.
Similarly, if we let $\alpha$ go to 0, we end up with MC.

It is easy to design toy environments where none of the obstacles described above apply, and where TD is the optimal choice (see Map 0 in Figure~\ref{fig:labyrinth_2d_value_functions_0_5}).
As a consequence of the switching and testing cost, it is not reasonable to expect our algorithm to dominate \emph{both} MC and TD in \emph{all} environments.
Our goal is to design an algorithm that is not worse than the worst of MC and TD in any scenario, and is close to the best one in most cases, or actually better.

\subsection{Formal Description}
Adaptive TD is presented as Algorithm~\ref{alg:batch_adaptive_td} in the appendix.
It has two components: confidence computation, and value function fitting.
First, we compute the MC target dataset $D_{\MC} = \{ (s, T_{MC}(s))_{s \in S} \}$ for all states $s$ in any input episode (we refer to $S$ as the union of those).
Then, we need to solve the regression problem with a method that provides confidence intervals: with probability $\alpha$ the expected return from $s$ under $\pi$ is in $(L_{MC}^{\alpha}(s), U_{MC}^{\alpha}(s))$.
There are a variety of approximate methods we can use to compute such confidence bounds; we discuss this in detail in the next subsection.
Note, however, that this can be regarded as an additional input or hyper-parameter to Adaptive TD.

In the second stage, after fixing a confidence function $\text{CI}_{MC}^{\alpha}(s)$, we apply a constrained version of TD.
We loop over all states $s \in S$ (possibly, in a randomized way, as the main loop over data can be replaced by a stochastic mini-batch), and we compute their TD target $T_{\TDz}(s)$. 
If the TD target falls within the confidence interval $\text{CI}_{MC}^{\alpha}(s)$, we simply use it to update $\wh{V}(s)$.
If it does not, i.e. when $T_{\TDz}(s) \notin (L_{MC}^{\alpha}(s), U_{MC}^{\alpha}(s))$, then we no longer trust the TD target, and use the mid-point of the MC interval, $(L+U)/2$.
We can also use the closest endpoint to $T_{\TDz}(s)$, either $L$ or $U$.

\subsection{Uncertainty Estimates}
Poor quality uncertainty estimates may \emph{severely} affect the performance of Adaptive TD.
In particular, in those states where the ground truth is not captured by the MC confidence intervals, the TD target will be \emph{forced} to bootstrap wrong values without any possibility of recovery.
In the case of neural networks, uncertainty estimation has been an active area of research in the last few years.
In some decision-making tasks, like exploration, all we need are \emph{samples} from the output distribution, while actual intervals are required for Adaptive TD.
A simple fix for models that provide samples is to take a number of them, and then construct an approximate interval. 

Common approaches include variational inference \cite{blundell2015weight}, dropout \cite{gal2016dropout}, Monte Carlo methods \cite{welling2011bayesian, mandt2016variational}, bootstrapped estimates \cite{efron1982jackknife, osband2016deep}, Gaussian Processes \cite{rasmussen2003gaussian}, or Bayesian linear regression on learned features \cite{snoek2015scalable, riquelme2018deep, azizzadenesheli2018efficient}.
While all the methods above could be used in combination with Adaptive TD, for simplicity, we decided to use an ensemble of $m$ MC networks \cite{lakshminarayanan2017simple}.
The algorithm works as follows.
We fit $m$ networks on the $D_{\MC} = \{ (s, T_{MC}(s))_{s \in S} \}$ dataset (we may or may not boostrap the data at the episode level).
Given a new state $s$, the networks provide value estimates $v_1, \dots, v_m$ at $s$.
We then compute a \emph{predictive} confidence interval, under the assumption that $v_i$ for $i=1, \dots, m$ are i.i.d. samples from some distribution $\mathbf{F}$.
Now, if $v_{m+1}$ was sampled from the same distribution, then we could expect $v_{m+1}$ to fall in the predictive interval with probability $\alpha$.
The idea is that the TD estimate should approximately correspond to another sample from the MC distribution.
If the deviation is too large, we will rely on the MC estimates instead.

In particular, we do assume $\mathbf{F}$ is Gaussian: $v_1, \dots, v_m \sim \mathcal{N}(\mu, \sigma^2)$ for unknown $\mu, \sigma^2$.
Let us define $\bar{v} = \sum_i v_i / m$, and $\hat{\sigma}_m^2 = \sum_i (v_i - \bar{v})^2 / (m-1)$.
Finally, if the assumptions hold, we expect that
\begin{equation}\label{eq:conf_intervals}
    \bar{v} - z_{\alpha} \hat{\sigma}_m \sqrt{1 + 1/m} \le v_{m+1} \le \bar{v} + z_{\alpha} \hat{\sigma}_m \sqrt{1 + 1/m} 
\end{equation}
with probability $\alpha$, where $z_{\alpha}$ is the $100(1-\alpha/2)$ percentile of the Student's distribution with $m-1$ degrees of freedom.
Then, we set $L_{MC}^{\alpha}(s)$ and $U_{MC}^{\alpha}(s)$ to the left and right-hand sides of \eqref{eq:conf_intervals} (note $v_i$ depends on $s$).
Of course, in practice the assumptions may not hold (for example, $v_i, v_j$ for $i \neq j$ will not be independent unless we condition on the data), but we still hope to get a reasonable estimate.

\section{A simple example}
\label{sec:simple_example}
In this section we illustrate the different bias-variance tradeoffs achieved by Monte Carlo and TD through a simple example, particularly highlighting the leakage propagation effect of TD described less formally in the previous section.

Consider the following MDP, with one initial state $s_0$ where  $k>1$ actions are available. Each action $a_i$ results in a deterministic transition to the corresponding state $s_i$. 
The first $p$ of these states $s_1,\dots,s_p$ then transfer the agent deterministically to state $b_1$, and the remaining states $s_{p+1},\dots,s_k$ are deterministically followed by $b_2$.
States $b_1$ and $b_2$ are followed by another deterministic transition to state $q$ and then to a final state, emitting a random $\mathcal{N}(\mu, \sigma^2)$ reward.
All other transitions lead to no reward.
The episode ends when the final state is reached. 

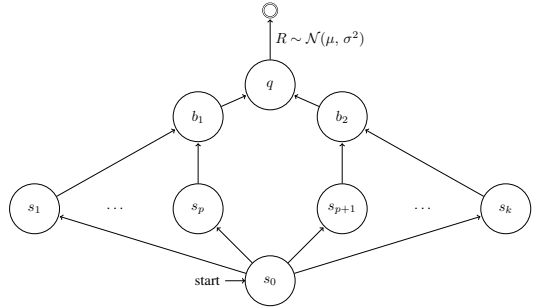
\begin{figure}[ht]
\centering
\resizebox{0.5 \columnwidth}{!}{%
\begin{tikzpicture}[state/.style={circle, draw, minimum size=1.2cm}]
]
    \node[initial, state]             (s0) {$s_0$};
    \node[draw=none, above=of s0] (empty) {};
    \node[state, left=of empty] (s3) {$s_p$};
    \node[draw=none, left=of s3] (s2) {$\cdots$};
    \node[state, left=of s2] (s1) {$s_1$};
    \node[state, right=of empty] (s4) {$s_{p+1}$};
    \node[draw=none, right=of s4] (s5) {$\cdots$};
    \node[state, right=of s5] (s6) {$s_k$};
    \node[state, above=of s3] (sB1) {$b_1$};
    \node[state, above=of s4] (sB2) {$b_2$};
    \node[draw=none, above=of empty] (empty2) {};
    \node[state, above=of empty2] (F) {$q$};
    \node[circle, draw, accepting, above=of F] (T) {};

    \draw[every loop]
        (s0) edge[right, auto=left]  node {} (s1)
        (s0) edge[right, auto=right] node {} (s3)
        (s0) edge[right, auto=right] node {} (s4)
        (s0) edge[right, auto=left]  node {} (s6)
        
        (s1) edge[right, auto=right] node {} (sB1)
        (s3) edge[right, auto=right] node {} (sB1)
        (s4) edge[right, auto=right] node {} (sB2)
        (s6) edge[right, auto=right] node {} (sB2)
        
        (sB1) edge[right, auto=right] node {} (F) 
        (sB2) edge[right, auto=right] node {} (F)
        (F) edge[right, auto=right] node {$R \sim \mathcal{N}(\mu,\,\sigma^{2})$} (T)
        
        ;        
\end{tikzpicture}
}%
\caption{Simple episodic Markov Decision Process.} \label{fig:markov_chain}
\end{figure}
Let $\pi$ be the uniform policy that, at $s_0$, chooses actions $a_{1:k}$ leading to $s_{1:k}$ with equal probability.
We assume the discount factor is $\gamma = 1$.
The true value function is $v_{\pi}(s) = \E\left[R\right] = \mu$ for all $s \in \mathcal{S}$. Suppose we need to estimate the value function from a set of $n$ episodes collected from $\pi$ with final rewards $r_1, \dots, r_n$.

Let us compare the performance of both the MC update \eqref{eq:mc_return} and the TD(0) target \eqref{eq:td0_target} when no function approximation is used, that is, when $\wh{V}$ can take any real value for any state.
Note all $n$ trajectories pass through state $q$, but on average only ${n}/{k}$ pass through each of the intermediate states $s_1, ..., s_k$.
The estimate for $q$ will be equal for MC and TD: $\wh{V}(q) = \sum_i r_i / n$.
Its distribution has mean $\E[\wh{V}(q)] = \E[R] = \mu$ and variance $\Var[\wh{V}(q)] = \sigma^{2}/{n}$.

On the other hand, the variance of $\wh{V}$ for states $s_{1:k}$ does differ significantly for TD and MC.
The variance of the MC estimate has variance $\Var[\wh{V}_{\MC}(s_i)] \approx k {\sigma^{2}} / {n}$, due to approximately\footnote{For clarity of exposition, we assume here that $n$ is large enough so that the random fluctuations of the number of visits around $n/k$ is negligible.} $n/k$ episodes going through each $s_{1:k}$.
However, the TD estimator correctly realizes that $\wh{V}(s_i) = \wh{V}(q)$, thus inheriting only the variance of the estimator in state $q$: $\Var[\wh{V}_{\MC}(s_i)] = {\sigma^{2}} / {n}$.
Thus, TD reduces the variance of the estimates by a factor of $k$. 

\begin{figure}[t]  
\begin{center}
\setkeys{Gin}{width=\linewidth}
\begin{tabularx}{\columnwidth}{XX}
\includegraphics{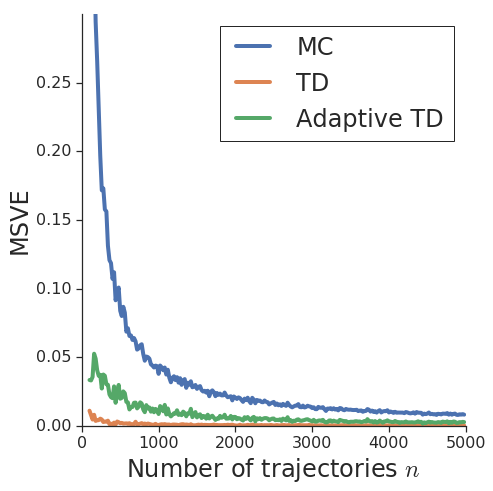} &
\includegraphics{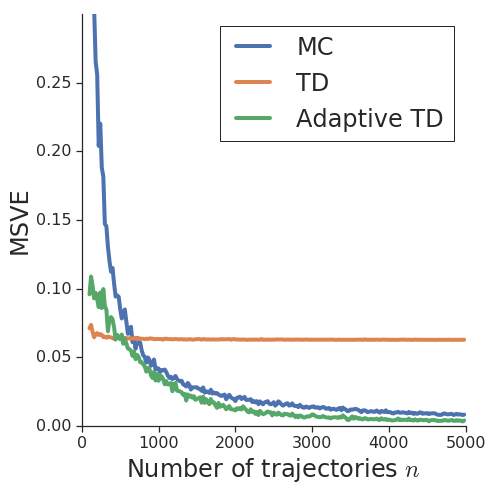}
\end{tabularx}
\caption{Mean-squared Value Error (MSVE) for MC, TD, and Adaptive TD as a function of the data size.
The left plot shows the tabular setting; the right plot shows the function approximation case with a fixed bias.
Each point is run 20 times to smooth out estimates, and $\mu = 0$. 
}
\label{fig:mc_vs_td}
\end{center}
\end{figure}

In the above setting with no function approximation, both methods are unbiased, and TD will dominate over MC for any finite size amount of data due to its reduced variance, while both converge to the same solution in the limit of infinite data.
However, assume now we use a function approximator that is \emph{not} able to correctly represent the value $\mu$ for either state $b_1$ or $b_2$, as it could happen in a general scenario with overly smooth function approximators.
TD estimates will introduce and propagate the bias to some of the states $s_1, ..., s_k$, while MC will not (noting though that for more general function approximators, MC estimates in some states may also influence estimates in others).
In this case, depending on the relative magnitude of TD's bias and MC's variance, and the amount of available data, one or the other will offer better performance. 
Figure \ref{fig:mc_vs_td} shows the performance of the two algorithms in the two settings, highlighting that the bias encoded in the function approximator impacts TD more severely than MC due to leakage propagation, when the number of sample episodes is high enough.
We report the MSVE for the intermediate $s_{1:k}$ states.
Notably, despite leakage, TD still outperforms MC when less data is available, due to its reduced variance.

\subsection{Results}
\begin{figure}[t]  
\begin{center}
\setkeys{Gin}{width=0.8 \linewidth}
\begin{tabularx}{0.5 \columnwidth}{XX}
\includegraphics{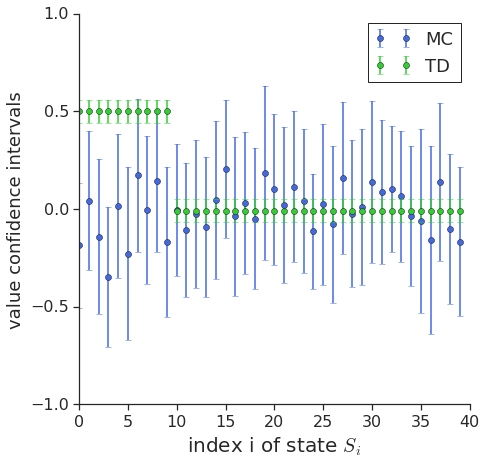}
\end{tabularx}
\caption{MC and TD estimates for the example of Section~\ref{sec:simple_example} ($\mu = 0$) with a biased function approximator.
Adaptive TD detects and corrects the mismatch between the TD updates and the MC confidence intervals.}
\label{fig:mc_vs_td2}
\end{center}
\end{figure}

We clearly see in Figure~\ref{fig:mc_vs_td} that Adaptive TD is able to take advantage from TD when the MC variance is too large (left), while avoiding bootstrapping erroneous values by detecting and correcting suspicious TD-estimates in the function approximation case (right).
Moreover, in Figure~\ref{fig:mc_vs_td2}, we see how in the latter case TD updates mostly fall outside the $95$\% Monte Carlo confidence intervals.
\section{Experimental Results}


\begin{figure*}[t]
\begin{center}
\setkeys{Gin}{width=\linewidth}
\begin{tabularx}{0.6 \columnwidth}{XX}
\includegraphics{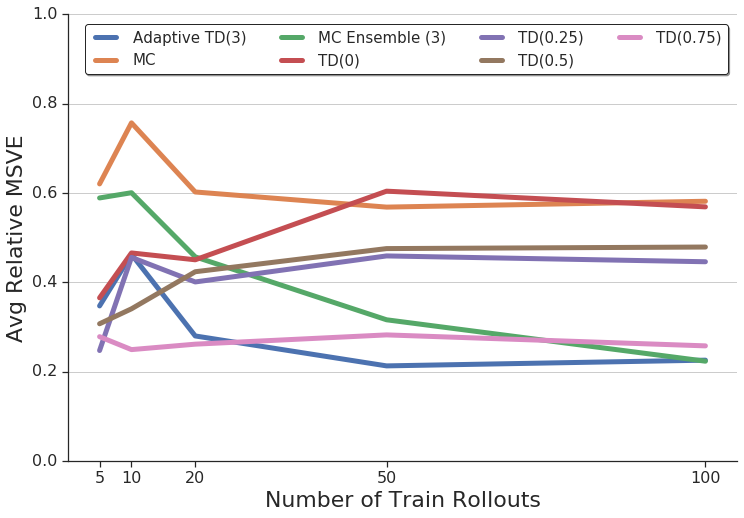}
\end{tabularx}
\caption{
Average normalized MSVE for Lab2D and Atari environments in each data regime.
For each number of train rollouts and scenario, we normalize the MSVE of each algorithm $A$ by $(\MSVE(A) - \min_{A^\prime} \MSVE(A^\prime)) / (\max_{A^\prime}\MSVE(A^\prime) - \min_{A^\prime}\MSVE(A^\prime))$. Equivalently, the worst algorithm is assigned relative MSVE 1, and the best one is assigned relative MSVE 0.
Then, for each number of rollouts, we take the average across scenarios (i.e., all the 10 environments are worth the same).
This allows for a reasonably fair comparison of performance in different domains.
}
\label{fig:overall_outcome}
\end{center}
\vspace*{-10pt}
\end{figure*}

In this section we test the performance of Adaptive TD in a number of scenarios that we describe below.
The scenarios (for which we fix a specific policy) capture a diverse set of aspects that are relevant to policy evaluation: low and high-dimensional state spaces, sharp value jumps or smoother epsilon-greedy behaviors, near-optimal and uniformly random policies.
We present here the results for Labyrinth-2D and Atari environments, and Mountain Car is presented in the appendix, Section \ref{app:mcar}.

We compare Adaptive TD with a few baselines: a single MC network, raw TD, and TD($\lambda$).
TD($\lambda$) is a temporal differences algorithm which computes an average of all $n$-step TD returns (an extension of the 1-step target in \eqref{eq:td0_target}), \cite{sutton1998introduction}.
For a clean comparison across algorithms in each scenario, we normalize the MSVE of all algorithms ($y$-axis) by the worst performing one, and we do this independently for each number of data rollouts ($x$-axis).
In addition, the appendix contains the absolute values with empirical confidence intervals for all cases.
Our implementation details are presented in Section~\ref{app:implementation} of the appendix.
In general, we did not make any effort to optimize hyper-parameters, as the goal was to come up with an algorithm that is robust and easy to tune across different scenarios.
Accordingly, for Adaptive TD, we use an ensemble of 3 networks trained with the MC target, and confidence intervals at the 95\% level.
The data for each network in the ensemble is bootstrapped at the rollout level (i.e., we randomly pick rollouts with replacement).
Plots also show the performance of the MC ensemble with 3 networks, to illustrate the benefits of Adaptive TD compared to its auxiliary networks.

\textbf{Labyrinth-2D.}
We first evaluate the performance of the algorithms in a toy scenario which represents a 2-d map with some target regions we would like to reach.
The state $s=(x, y)$ are the coordinates in the map, and the policy takes a uniformly random angle and then applies a fixed-size step.
The initial state $s_0$ for each episode is selected uniformly at random inside the map, and the episode ends after each step with probability $p = 0.0005$.
Reward is $r = 30$ inside the green regions, $r = 0$ elsewhere.
The maps layouts and their value functions are shown in Figure~\ref{fig:labyrinth_2d_value_functions_0_5} in the appendix.
The simple different layouts cover a number of challenging features for policy evaluation: sharp jumps in value near targets, several kind of walls, and locked areas with no reward (see maps 2 and 3). Due to the randomized policy and initial state, we tend to uniformly cover the state space.
We run experiments with $n = 5, 10, 20, 50, 75, 100$ training episodes.
We approximate the ground truth in a grid by sampling and averaging a large number of test episodes from each state.

\begin{figure*}[t]
\setkeys{Gin}{width=\linewidth}
\begin{tabularx}{\columnwidth}{XX}
\includegraphics{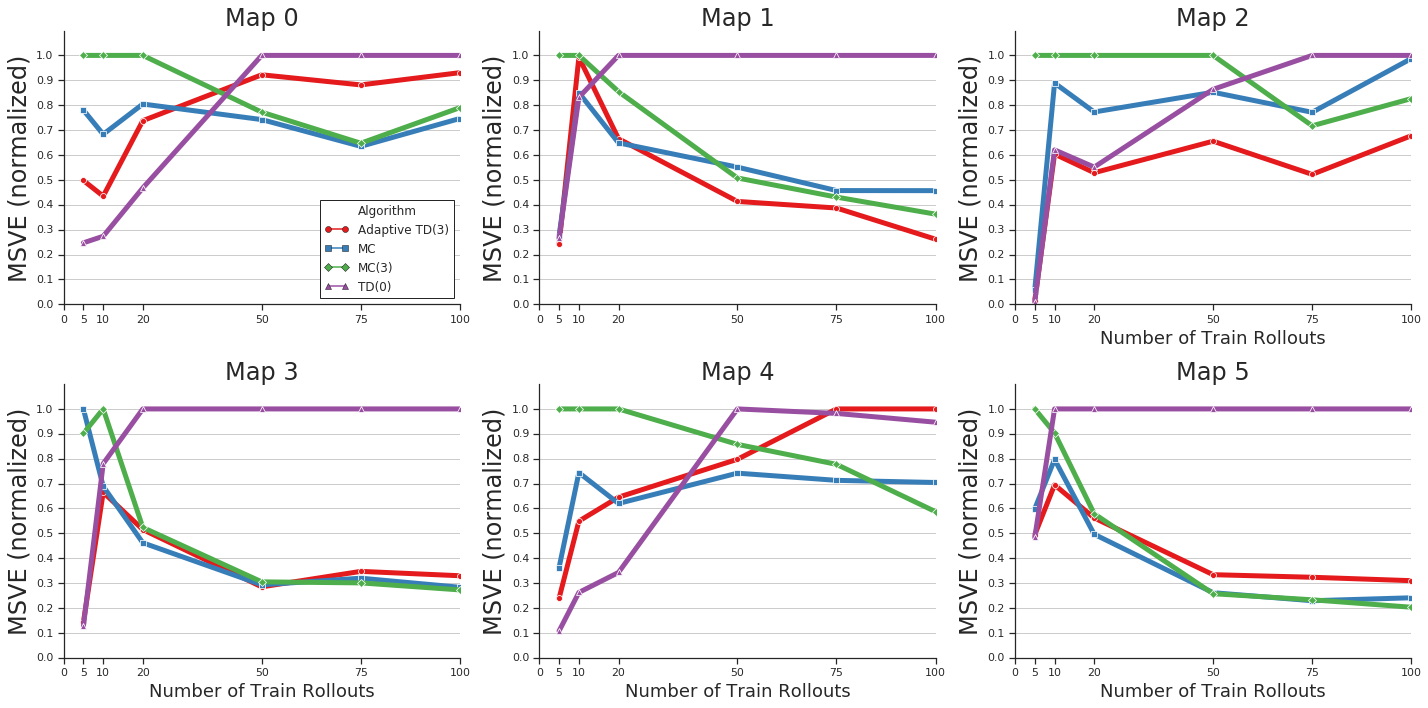}
\end{tabularx}
\caption{
\textbf{Labyrinth-2D.}
For each number of train rollouts, we normalize the MSVE of each algorithm $A$ by $\MSVE(A) / \max_{A^\prime}\MSVE(A^\prime)$.
Absolute numbers and conf.\ intervals in the appendix.}
\label{fig:lab2_outcome}
\end{figure*}

The results are shown in Figure~\ref{fig:lab2_outcome}.
As expected, in most of the maps we observe that MC outperforms TD in high-data regimes, while MC consistently suffers when the number of available data rollouts is limited.
Adaptive TD shows a remarkably robust performance in all cases, being able to strongly benefit from TD steps in the low-data regime (see, basically, all maps) while remaining very close to MC's performance when a large number of rollouts are available.
In that regime, the improvement with respect to TD is dramatic in maps that are prone to strong leakage effects, like maps 1, 2, 3, and 5.
In Figure~\ref{fig:lab2_maps_lambdas_unnormalized} in the appendix, we can also see the results for TD($\lambda$).
In this particular case, it seems $\lambda = 0.75$ is a good choice, and it is competitive with Adaptive TD in challenging maps 1, 2, 3, and 5.
However, the remaining values of $\lambda$ are mostly outperformed in these scenarios.
Figure \ref{fig:map2_conf_int_violations} shows the regions of the maps state space where the TD target falls outside the MC interval for Adaptive TD.

\textbf{Atari.}
The previous examples illustrate many of the practical issues that arise in policy evaluation.
In order to model those issues in a clean disentangled way, and provide some intuition, we focused so far on lower-dimensional state spaces. 
In this section we evaluate all the methods in a few Atari environments \cite{bellemare2013arcade}: namely, Breakout, Space Invaders, Pong, and MsPacman.
The state consists of four stacked frames, each with $(84, 84)$ pixels, and the initial one is fixed for each game.
We would like to focus on competitive policies for the games, while still offering some stochasticity to create a diverse set of trajectories (as Atari environments are deterministic).
We use soft Q policies that sample from the action distribution to generate the training and test trajectories.
The temperature of the softmax layer was adjusted to keep a good trade-off between trajectory diversity and performance of the policy.
Directly computing the ground-truth value is not feasible this time, so we rely on a large number of test rollouts to evaluate our predictions.
This increases the variance of our MSVE results.

\begin{figure*}[!ht]
\setkeys{Gin}{width=\linewidth}
\begin{tabularx}{\columnwidth}{XX}
\includegraphics{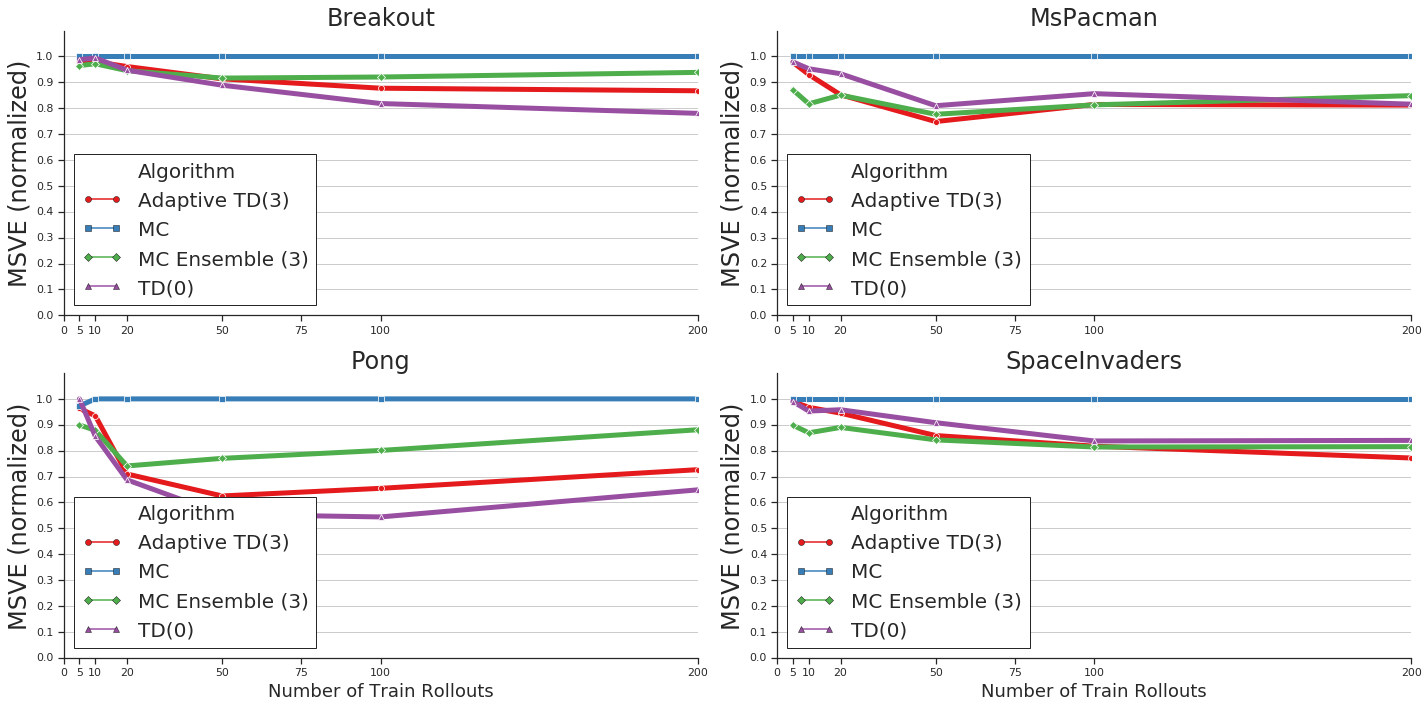}
\end{tabularx}
\caption{
\textbf{Atari.}
For each number of train rollouts, we normalize the MSVE of each algorithm $A$ by $\MSVE(A) / \max_{A^\prime}\MSVE(A^\prime)$.
Absolute numbers and confidence intervals are in the appendix.
}
\label{fig:atari_outcome}
\vspace*{-10pt}
\end{figure*}

The results are shown in Figure~\ref{fig:atari_outcome}.
TD does a good job for all number of rollouts.
This suggests that in high-dimensional state spaces (like Atari frames) the required number of samples for MC to dominate may be extremely large.
In addition, a single MC network seems to struggle in all games, while the prediction of the MC ensemble proves significantly more robust.
Adaptive TD outperforms MC, and its auxiliary MC ensemble.
Moreover, it offers a performance close to that of TD, maybe due to wide MC confidence intervals in high-dimensional states which reduce Adaptive TD to simply TD in most of the states. 
We show the results for TD($\lambda$) in Figure~\ref{fig:atari_outcome_app} in the appendix.
In this case, $\lambda = 0.75$ --which did a good job in Labyrinth-2D scenarios-- is always the worst.
In particular, Adaptive TD improvements compared to TD($\lambda = 0.75$) range from 30\% in the low-data regimes of Pong, to consistent 20\% improvements across all data regimes of Space Invaders.

\textbf{Summary.} Figure~\ref{fig:overall_outcome} displays the overall results normalized and averaged over the 10 scenarios.
Adaptive TD strongly outperforms TD and MC, and offers significant benefits with respect to its auxiliary ensemble, and TD($\lambda$).
This highlights the main feature of Adaptive TD: its \emph{robustness}.
While TD and MC outperform each other often by a huge margin depending on the scenario and data size, Adaptive TD tends to automatically mimic the behavior of the best-performing one.
TD($\lambda$) methods offer a way to interpolate between TD and MC, but they require to know a good value of $\lambda$ in advance, and we have seen that this value can significantly change across problems.
In most cases, Adaptive TD was able to perform --at least-- comparably to TD($\lambda$) for the \emph{best} problem-dependent $\lambda$.


\section{Related Work}
Both $n$-step TD and TD($\lambda$) offer practical mechanisms to control the balance between bias and variance by tuning $n$ and $\lambda$. However, these parameters are usually set {\sl a priori}, without taking into account the progress of the learning process, and are not state-dependant.   

A number of works have addressed on-policy evaluation.
In \cite{mann2016adaptive} the authors introduce an algorithm for batch on-policy evaluation, which is capable of selecting the best $\lambda$ parameter for LSTD by doing efficient cross-validation. However, it only works for the case of linear function approximation, and does not take per-state decisions. 
In the TD-BMA algorithm \cite{downey2010temporal}, decisions are taken by state, like in ours, while TD-BMA is restricted to the tabular setting. We cover function approximation, including deep neural networks, and large dimensional input spaces, where uncertainty estimates require different techniques.


Per-state $\lambda$ was used for the off-policy case in \cite{Sutton2014Q}. (For an overview of methods with fixed $\lambda$, including the off-policy case, see \cite{Geist2014}.) This study is continued in \cite{Sutton2015} with methods that selectively (de-)emphasize states. The motivation is that function approximation has to give a compromise function (e.g. forcing similar values at nearby states, even if the observed values are not similar), which should be guided by emphasizing the more important states.
In off-policy evaluation (the focus of their paper) there is more interest in states that will occur more frequently in our actual policy. 
Similarly, when we switch to MC for some states in our approach, we may be less interested to model their value correctly in the TD
function approximation.
This paper also hints that $\lambda(s)$ may be modified depending on the variance of the returns after $s$; this is then developed in \cite{White2016AGA}.

The algorithm of \cite{White2016AGA} estimates the variance of the $\lambda$--returns arising from the data by establishing a Bellman operator for the squared return, for which they are looking for a fixed point. Then they optimize ``greedily'' the $\lambda(s_t)$ such that they get the optimal bias/variance trade-off for this state. However, their variance of the returns is restricted
to the uncertainty coming from the actions and returns, but does not take into account the model uncertainty arising from the function approximation (which we include here by evaluating an ensemble of networks).
\cite{Thomas2015} introduced a different approach to compute an optimized and state-dependent combination of $n$--step--returns. 
They ask what the optimal
combination of $n$--step returns would be if the estimates were
unbiased and their variances and covariances were known. 
This differs from our approach as we are also trying to minimize the bias that is introduced by function approximation and
that is amplified by TD's bootstrapping.

\section{Future Work}
There are a number of avenues for future work.
Adaptive TD can be easily extended to $n$-step TD or TD$(\lambda)$: at each transition, the agent checks whether the potential $n$-step or TD$(\lambda)$ target is within the associated confidence interval. If so, the agent applies the TD update; otherwise, it proceeds to the next transition and carries over the unused target to compute the next TD target and repeat the process.
In addition, the proposed policy evaluation algorithm can be implicitly or explicitly incorporated into a policy improvement one for control.
Finally, we expect that constructing more sophisticated and accurate confidence intervals based on MC returns will improve the performance of Adaptive TD.

\clearpage

\bibliographystyle{plain}
\bibliography{references}

\clearpage

\appendix
\clearpage
\section{Implementation Details and Neural Network Hyper-Parameters}
\label{app:implementation}

Our experiments in the Labyrinth-2D and Mountain Car environments, which both have 2d state spaces, were conducted using a multi-layer perceptron with exactly the same configuration. For Atari, we followed the standard pre-processing from the DQN paper \cite{mnih2015human}, where the inputs are reduced to 84x84, and 4 consecutive frames are stacked. The architecture of the convnet used is also standard, with the exception that instead of 18 outputs, we only have 1, as we are estimating state-value functions, not action-value functions.
Details are provided in Table \ref{table:nets_and_optimizers} below.

The implementation of TD does \emph{not} use a target network; we consider the current estimates for the target (while we do not optimize with respect to the target, i.e., we apply a stop-gradient operation).
In general, we did not make any effort to optimize hyper-parameters.

Table \ref{table:nets_and_optimizers} summarizes the parameters of the neural networks used as value function approximators and the training hyper-parameters used in the experiments on the 2D environments and ATARI. The parameters for ATARI are the same as the ones from the original DQN paper \cite{mnih2015human}.

\begin{table}[h]
\begin{center}
\begin{tabular}{|l|l|l|}
  \hline
  \multicolumn{3}{|l|}{\textbf{Neural network}}  \\
  \hline \hline
   & 2D Envs. & Atari \\
  \hline
  Input dimensions & 2 &  84 x 84 x 4 \\ 
  Convnet output channels & - & (32, 64, 64)\\
  Convnet kernel shapes & - & (8, 4, 3)\\
  Convnet strides & - & (4, 2, 1)\\
  Linear layers hidden-units & (50, 50) & (512) \\
  Non-linearities & Relu & Relu\\
  \hline
  \hline
  \multicolumn{3}{|l|}{\textbf{Training hyper-parameters}}  \\
  \hline
  \hline
  Mini-batch size & 512 & 32 \\
  Optimiser algorithm & Adam & Adam\\ 
  Learning rate & 0.001 & 0.0000625 \\
  Beta1 & 0.9 & 0.9 \\
  Beta2 & 0.999 & 0.999 \\
  Epsilon & 1e-08 & 0.00015 \\
  Training batches & 50000 & 250000 \\
  \hline
\end{tabular}
\caption{\label{table:nets_and_optimizers}Neural network architectures and hyper-parameter details.}
\end{center}
\end{table}

\subsection{Algorithm Pseudo-code}

We present the pseudo-code for Adaptive TD as Algorithm~\ref{alg:batch_adaptive_td}.

\begin{algorithm}[tb]
  \caption{Adaptive TD}
  \label{alg:batch_adaptive_td}
\begin{algorithmic}
  \STATE {\bfseries Input:} Confidence level $\alpha \in (0, 1)$. 
  Trajectories $\tau_{1}, \dots, \tau_n$ generated by policy $\pi$.
   \\
  \hrulefill
  \STATE Let $S$ be the set of visited states in $\tau_{1}, \dots, \tau_n$. Initialize $\wh{V}(s) = 0$, for all $s$.
  \STATE Compute Monte-Carlo returns dataset as in \eqref{eq:mc_return}:
  $D_{\MC} = \{ (s, T_{MC}(s))_{s \in S} \}.$
  \STATE Fit confidence function to $D_{\MC}$:
  $\text{CI}_{MC}^{\alpha}(s) := (L_{MC}^{\alpha}(s), U_{MC}^{\alpha}(s)).$
  \REPEAT
      \FOR{$i=1$ {\bfseries to} $n$}
          \FOR{$t=1$ {\bfseries to} $|\tau_i| - 1$}
            \STATE {$s_{t}^{(i)}$ is the $t$-th state of $\tau_i$.}
            \STATE {$T_{\TDz} = r(s_{t}^{(i)}) + \gamma \ \wh{V} ( s_{t+1}^{(i)})$} 
            \IF{$T_{\TDz} \in (L_{MC}^{\alpha}(s_{t}^{(i)}), U_{MC}^{\alpha}(s_{t}^{(i)}))$ } 
                \STATE $T_{i, t} \leftarrow T_{\TDz}$
            \ELSE 
                \STATE $T_{i, t} \leftarrow (L_{MC}(s_{t}^{(i)}) +  U_{MC}(s_{t}^{(i)}) / 2$
            \ENDIF
            \STATE Use target $T_{i, t}$ to fit $\wh{V}(s_{t}^{(i)})$.
          \ENDFOR
      \ENDFOR
  \UNTIL{epochs exceeded} \\
  \hrulefill
\end{algorithmic}
\end{algorithm}

\subsection{Online Scenarios}
For policy evaluation, we assume all data is collected in advance.
Sometimes this assumption may be too strong, and we would also like to allow for updates in our confidence intervals based on a stream of new data.
When the data comes from the same policy, the extension should be straightforward.
In general, we can train both estimates in parallel (say, an MC ensemble, and a TD network), and freeze both of them every fixed number of updates as it is nowadays standard in DQN (target network) to keep training a copy of the TD network.
We leave the exploration of these extensions as future work.
\clearpage
\section{Mountain Car Environment}
\label{app:mcar}

We also test the algorithms in the popular Mountain Car environment, where the goal is to control a car in order to climb a steep hill.
The state has two coordinates, corresponding to the velocity and position of the car, and there are three actions: move left, move right, and do nothing.
We use a near-optimal policy together with $\epsilon$-greedy steps, for $\epsilon=0.2$.
We compute the ground truth value function as in the previous case, see Figure~\ref{fig:mountain_car_value_function} below.

\begin{figure}[!ht]
\begin{center}
\setkeys{Gin}{width=\linewidth}
\begin{tabularx}{0.5 \columnwidth}{X}
\includegraphics{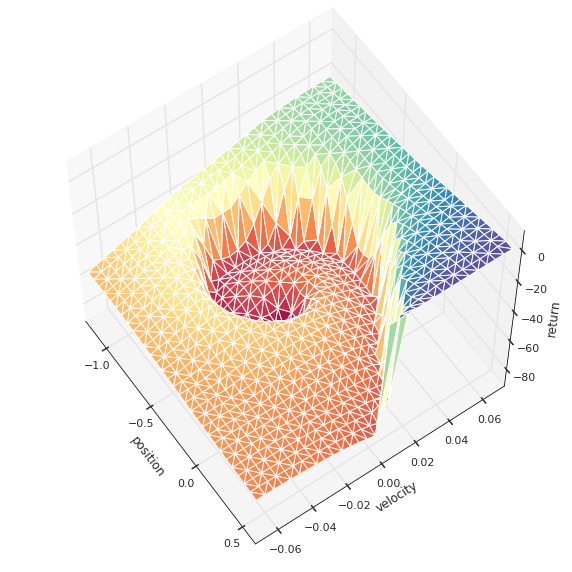}
\end{tabularx}
\caption{Mountain Car value function for near-optimal policy with epsilon greedy ($\epsilon=0.2$) actions.}
\label{fig:mountain_car_value_function}
\end{center}
\end{figure}

\begin{figure}[ht] 
\begin{center}
\setkeys{Gin}{width= \linewidth}
\begin{tabularx}{\columnwidth}{XX}
\includegraphics{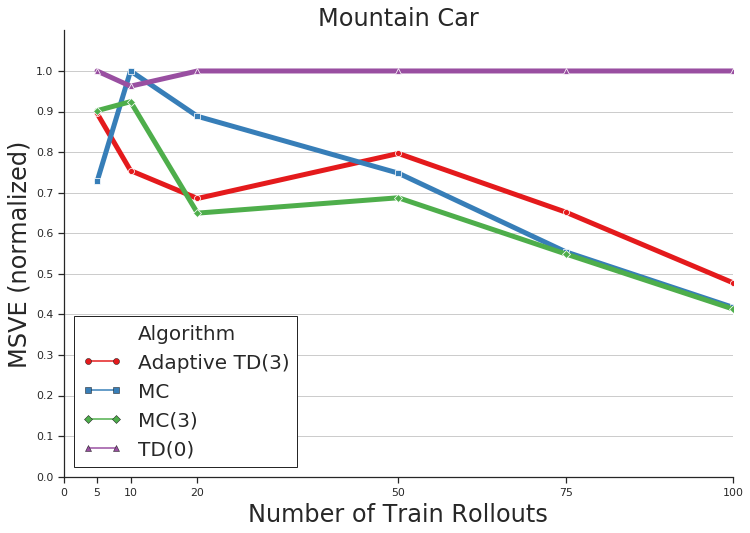} &
\includegraphics{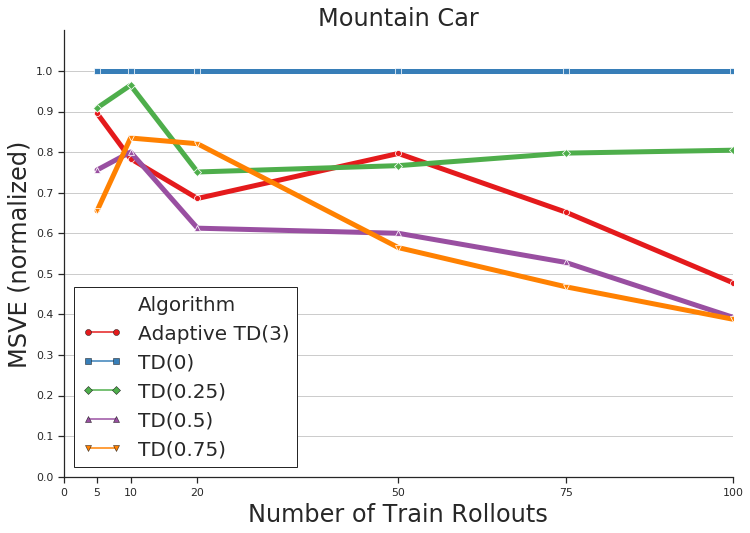}
\end{tabularx}
\caption{
\textbf{Mountain Car.}
For each number of train rollouts, we normalize the MSVE of each algorithm $A$ by $\MSVE(A) / \max_{A^\prime}\MSVE(A^\prime)$.}
\label{fig:mountaincar_outcome}
\end{center}
\end{figure}

The results are shown in Figure~\ref{fig:mountaincar_outcome} (left).
In this case MC seems to consistently outperform TD.
Adaptive TD offers strong performance in the low-data regime, and it mimics the behavior of MC when more data is available (and, presumably, we have access to better confidence intervals).
The gains with respect to TD are significant.
Figure~\ref{fig:mountaincar_outcome} (right) shows TD($\lambda$) methods with $\lambda \ge 0.5$ are competitive with the MC ensemble and with Adaptive TD.

\begin{figure}[t!]
\begin{center}
\setkeys{Gin}{width=\linewidth}
\begin{tabularx}{\columnwidth}{XX}
\includegraphics{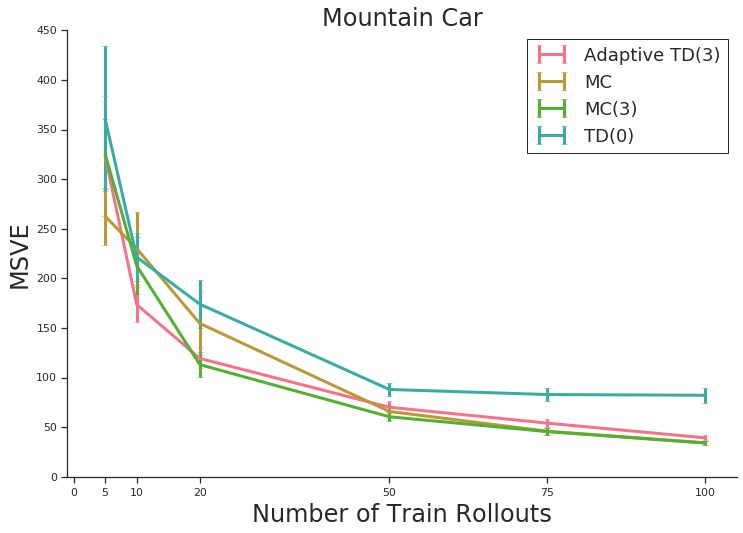} &
\includegraphics{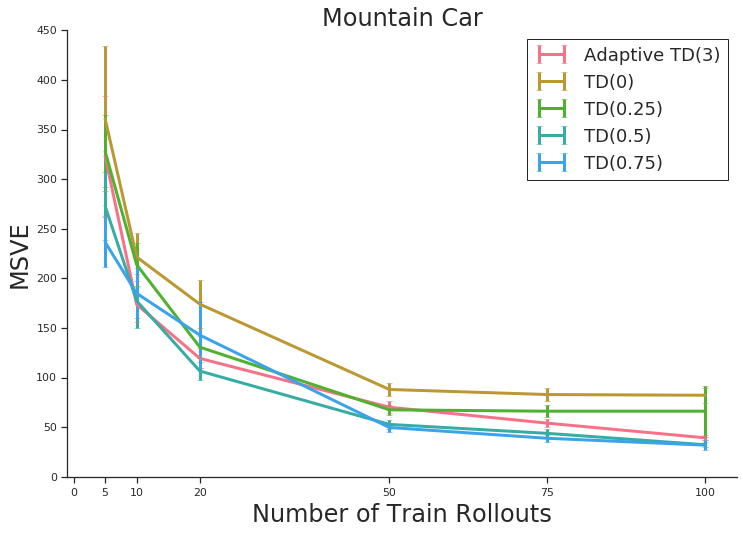}
\end{tabularx}
\caption{
MSVE for the Mountain Car environment.
Confidence intervals over 20 seeds.
}
\label{fig:mountaincar_outcome_unnormalized}
\end{center}
\end{figure}

Finally, for completeness, we show in Figure~\ref{fig:mountaincar_outcome_unnormalized} the unnormalized version of the plots in Figure~\ref{fig:mountaincar_outcome}.

\clearpage
\section{Labyrinth-2D Environments}

\begin{figure*}[h]
\setkeys{Gin}{width=\linewidth}
\begin{tabularx}{\columnwidth}{XX|XX}
\includegraphics{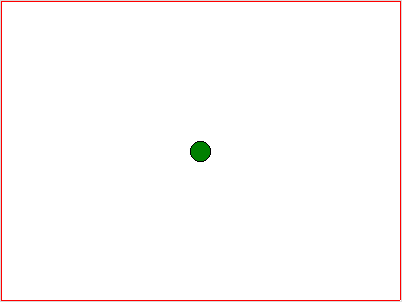} &
\includegraphics{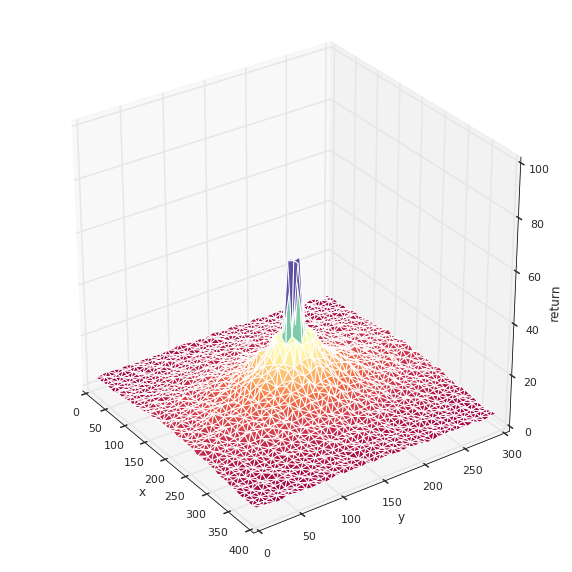} &
\includegraphics{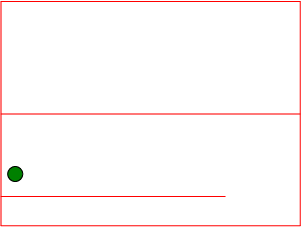} &
\includegraphics{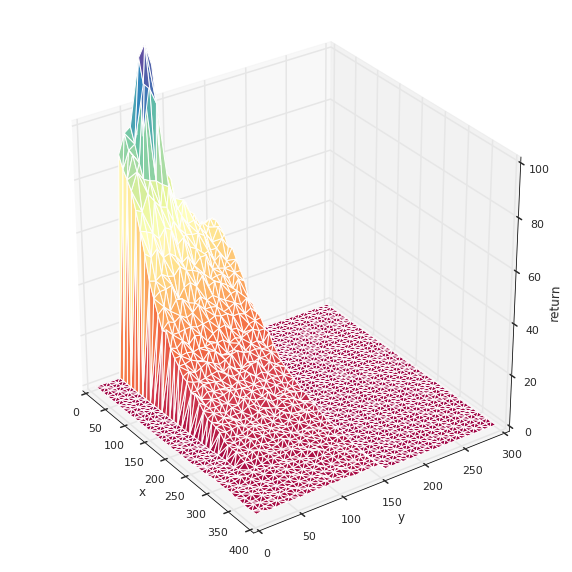} \\

\includegraphics{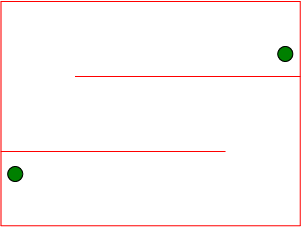} &
\includegraphics{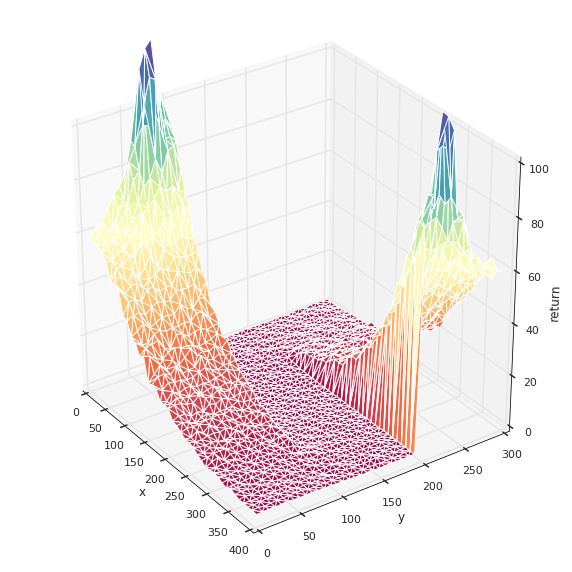} &
\includegraphics{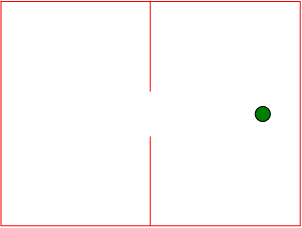} &
\includegraphics{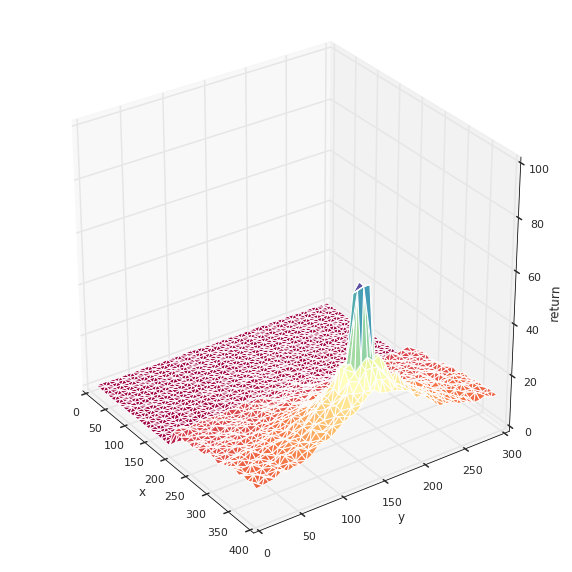} \\

\includegraphics{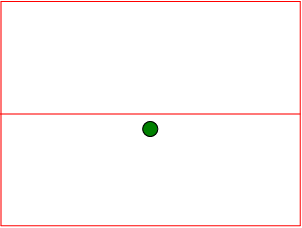} &
\includegraphics{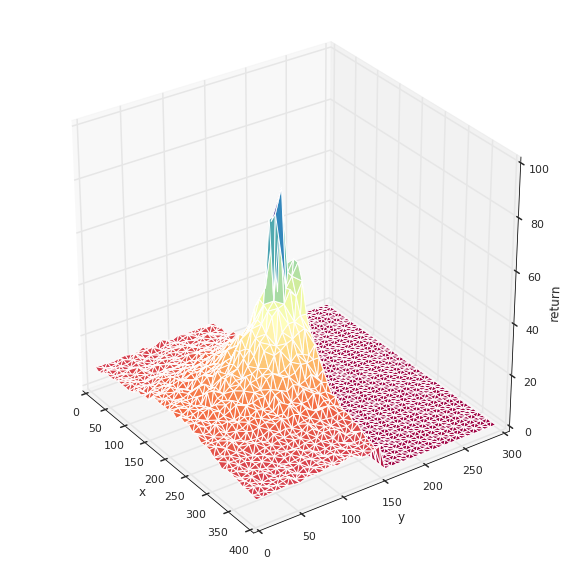} &
\includegraphics{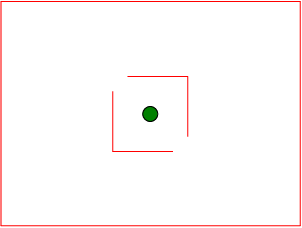} &
\includegraphics{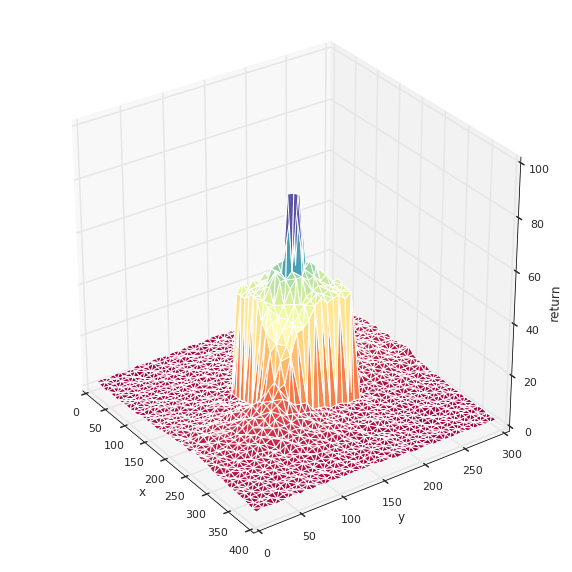} \\

\end{tabularx}
\caption{True value functions for the uniformly random policy in Labyrinth-2D: Maps 0, 1 and 2 (Left), Maps 3, 4, 5 (Right).}
\label{fig:labyrinth_2d_value_functions_0_5}
\end{figure*}

\begin{figure*}[h]
\setkeys{Gin}{width=\linewidth}
\begin{tabularx}{\columnwidth}{XX}
\includegraphics{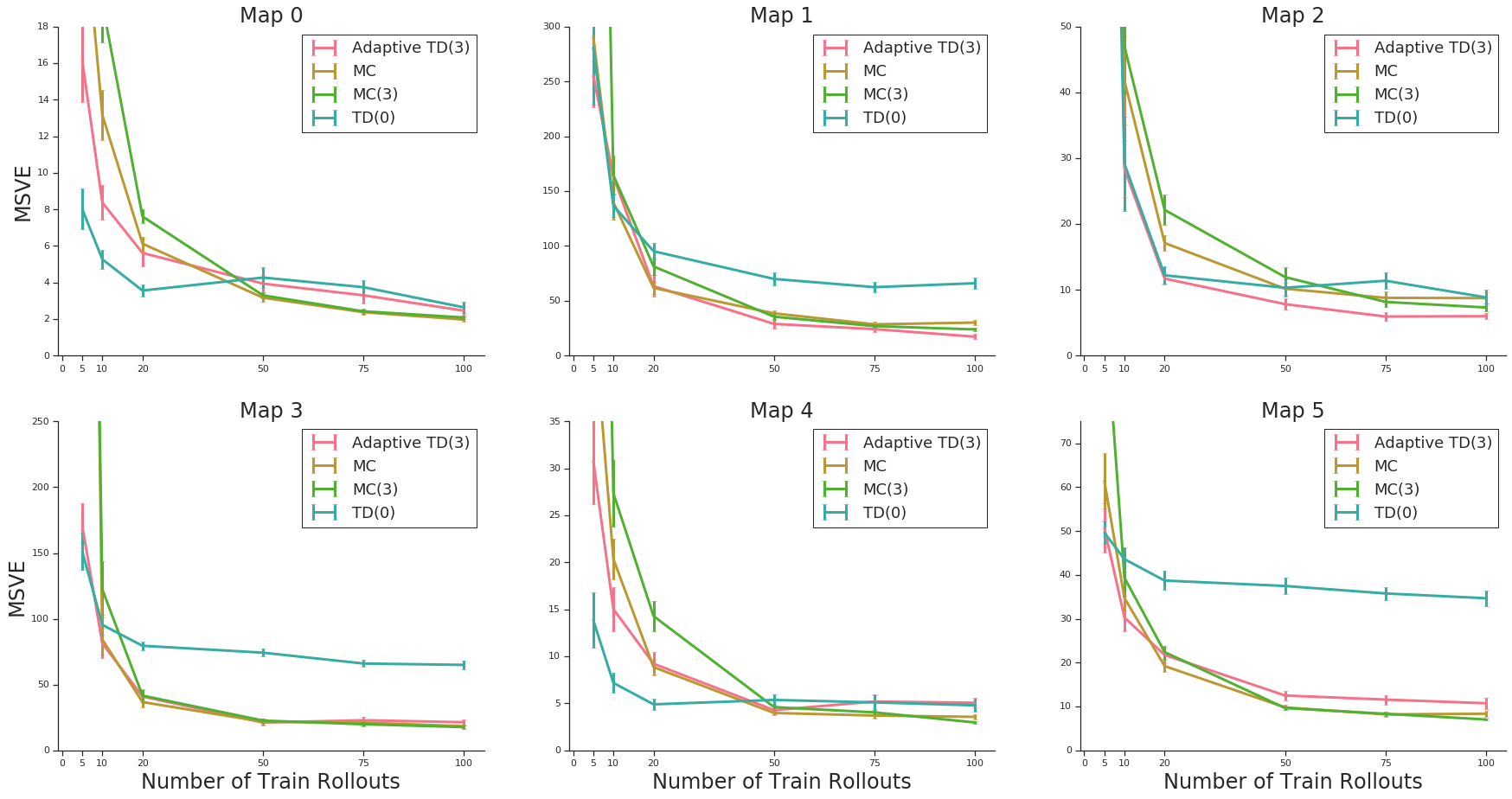}
\end{tabularx}
\caption{
\textbf{Labyrinth-2D.}
Unnormalized MSVE results.
Confidence intervals over 20 seeds.
}
\label{fig:lab2_maps_baselines_unnormalized}
\end{figure*}

\begin{figure*}[h]
\setkeys{Gin}{width=\linewidth}
\begin{tabularx}{\columnwidth}{XX|XX}
\includegraphics[width=3cm,height=2.25cm]{labyrinth_2d/map_0.png} &
\includegraphics[width=3.5cm,height=2.25cm]{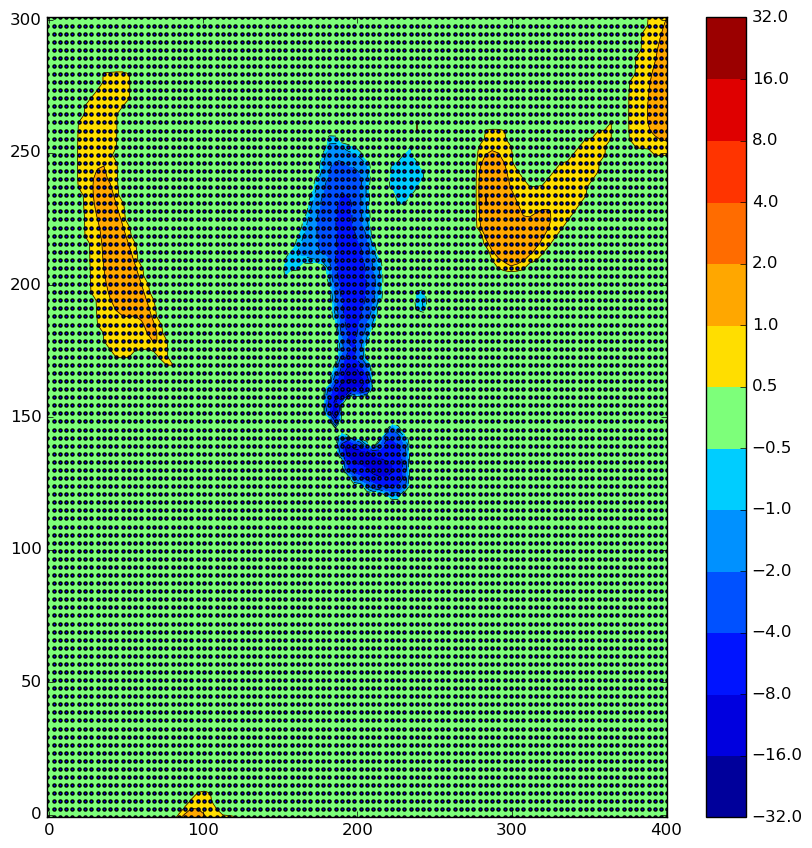} &
\includegraphics[width=3cm,height=2.25cm]{labyrinth_2d/map_3.png} &
\includegraphics[width=3.5cm,height=2.25cm]{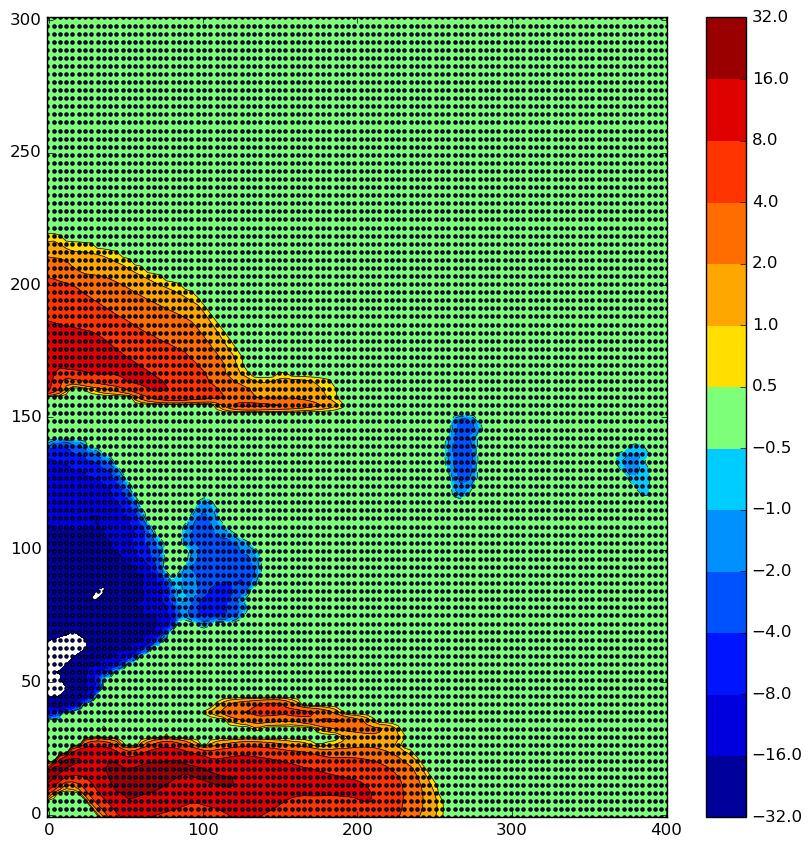} \\

\includegraphics[width=3cm,height=2.25cm]{labyrinth_2d/map_1.png} &
\includegraphics[width=3.5cm,height=2.25cm]{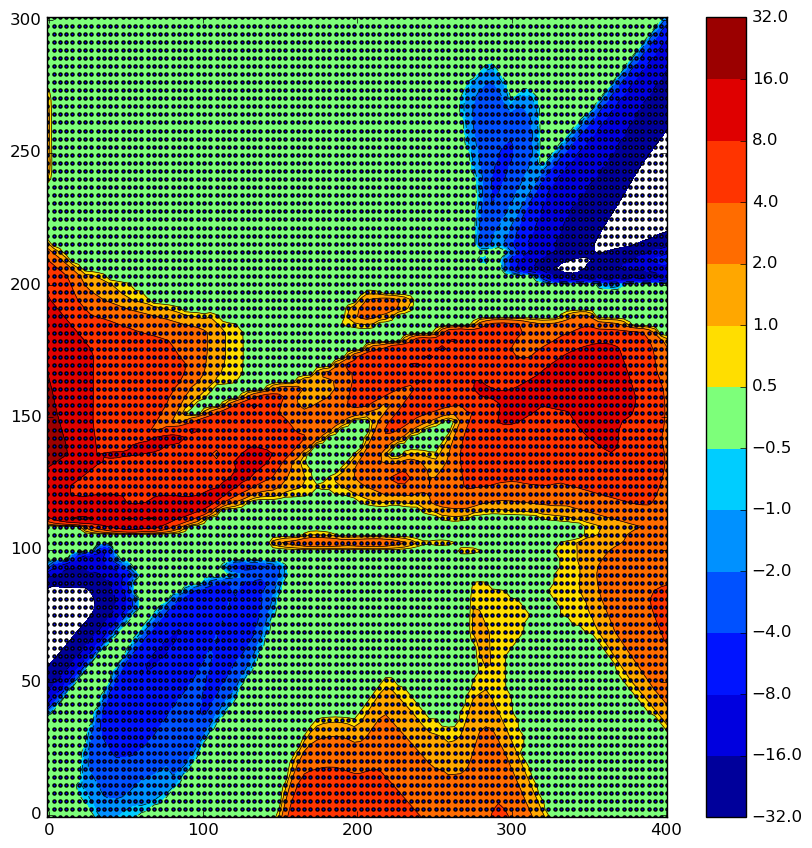} &
\includegraphics[width=3cm,height=2.25cm]{labyrinth_2d/map_4.png} &
\includegraphics[width=3.5cm,height=2.25cm]{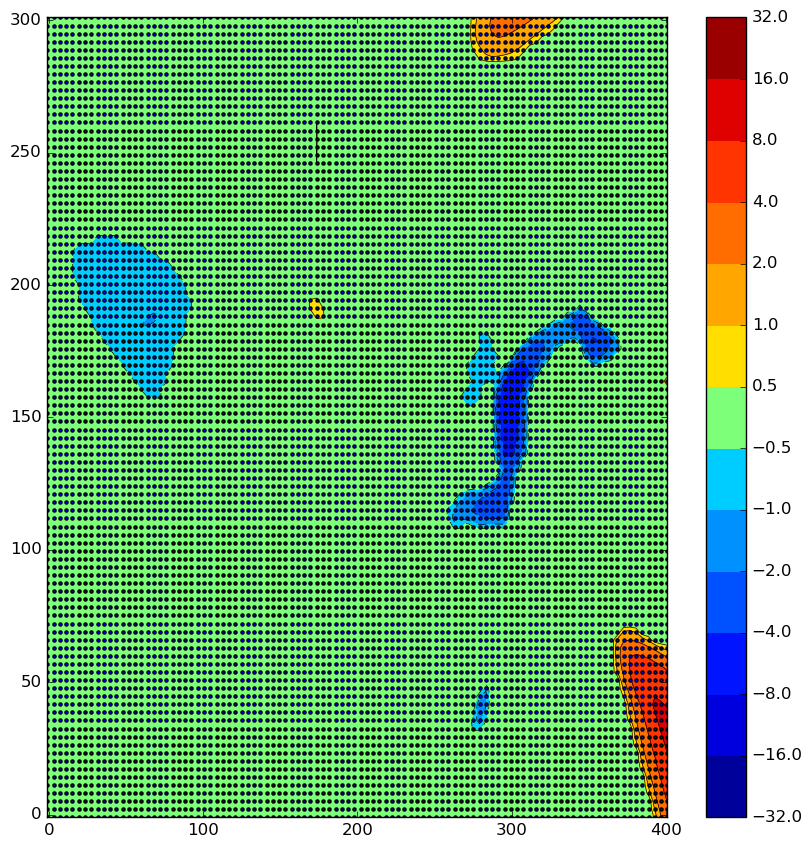} \\

\includegraphics[width=3cm,height=2.25cm]{labyrinth_2d/map_2.png} &
\includegraphics[width=3.5cm,height=2.25cm]{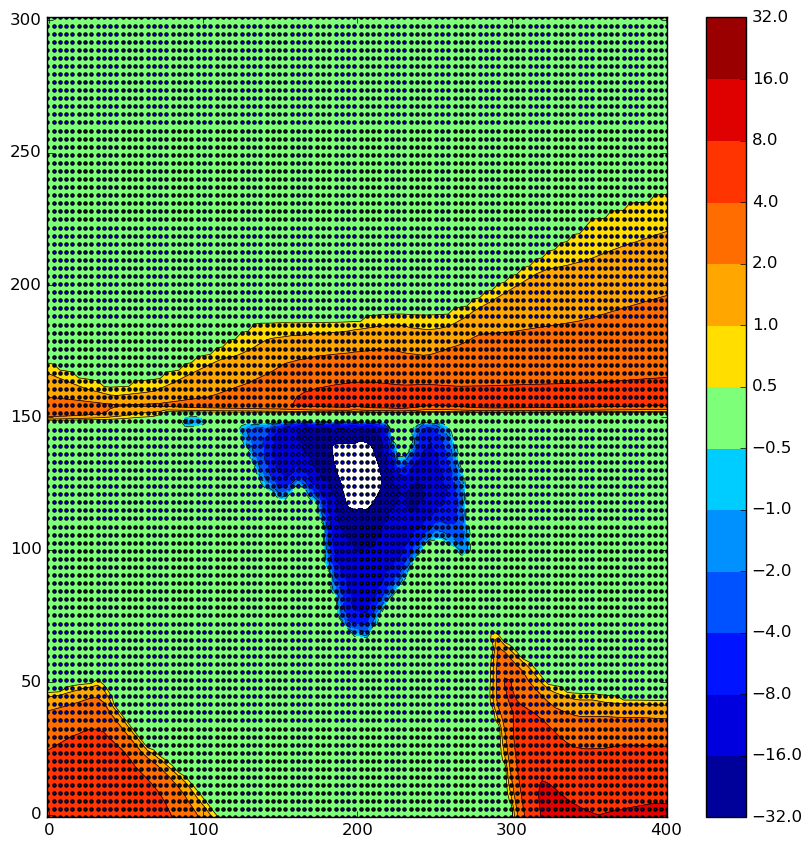} &
\includegraphics[width=3cm,height=2.25cm]{labyrinth_2d/map_5.png} &
\includegraphics[width=3.5cm,height=2.25cm]{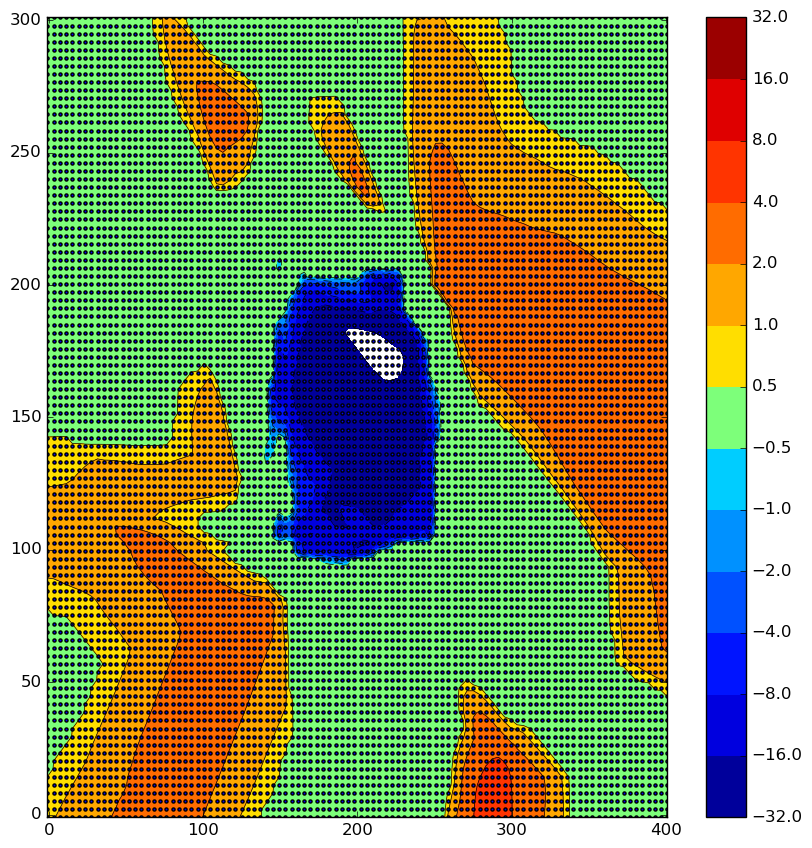} \\

\end{tabularx}
\caption{Adaptive TD confidence interval violations per state in the Labyrinth 2D map layouts, after 300 iterations of training, in the regime with 50 training rollouts. Green means the TD estimate is inside the MC confidence interval for that state. Red regions are over-estimations and blue regions under-estimations.
We see that corrections are needed in significant regions of the space at this stage.}
\label{fig:map2_conf_int_violations}
\end{figure*}

We created a set of six 2D maps used as toy environments in this paper. The layout of those environments always includes at least one reward represented as a green disk on Figure~\ref{fig:labyrinth_2d_value_functions_0_5}. The ground truth value function of a random policy navigating in those environments is shown on the same Figure.
This ground truth is used as a reference to compute the MSVE of the adaptive TD algorithm as well as the other baselines (Figure \ref{fig:lab2_maps_baselines_unnormalized}). In Figure~\ref{fig:map2_conf_int_violations}, we also present some insights on the TD versus MC decisions of the adaptive TD algorithm. We can clearly see that TD is selected for most of the states but the ones next to the wall where MC is preferred to prevent further leaking of the approximation errors observed near the wall.

\begin{figure*}[h]
\setkeys{Gin}{width=\linewidth}
\begin{tabularx}{\columnwidth}{XX}
\includegraphics{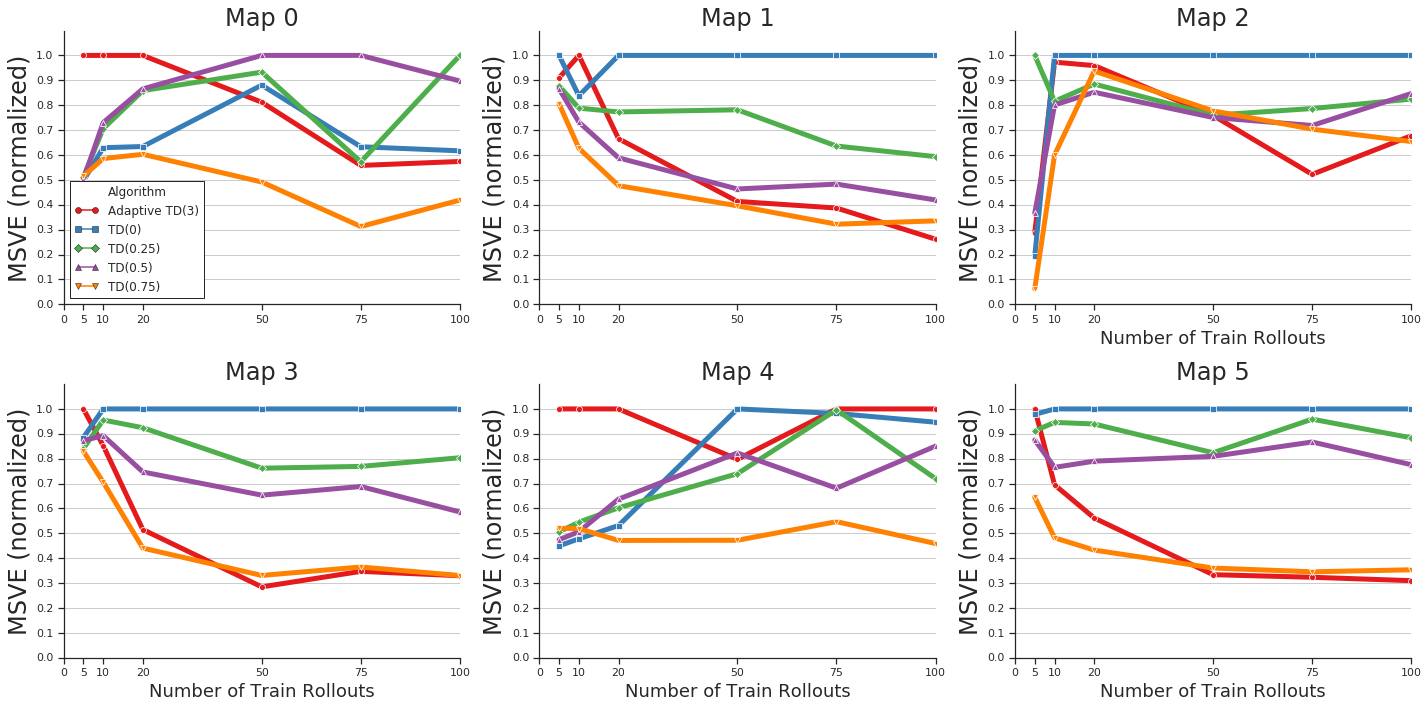}
\end{tabularx}
\caption{
\textbf{Labyrinth-2D.}
TD($\lambda$) results.
For each number of train rollouts, we normalize the MSVE of each algorithm $A$ by $\MSVE(A) / \max_{A^\prime}\MSVE(A^\prime)$.
}
\label{fig:lab2_maps_lambda}
\end{figure*}

\begin{figure*}[h]
\setkeys{Gin}{width=\linewidth}
\begin{tabularx}{\columnwidth}{XX}
\includegraphics{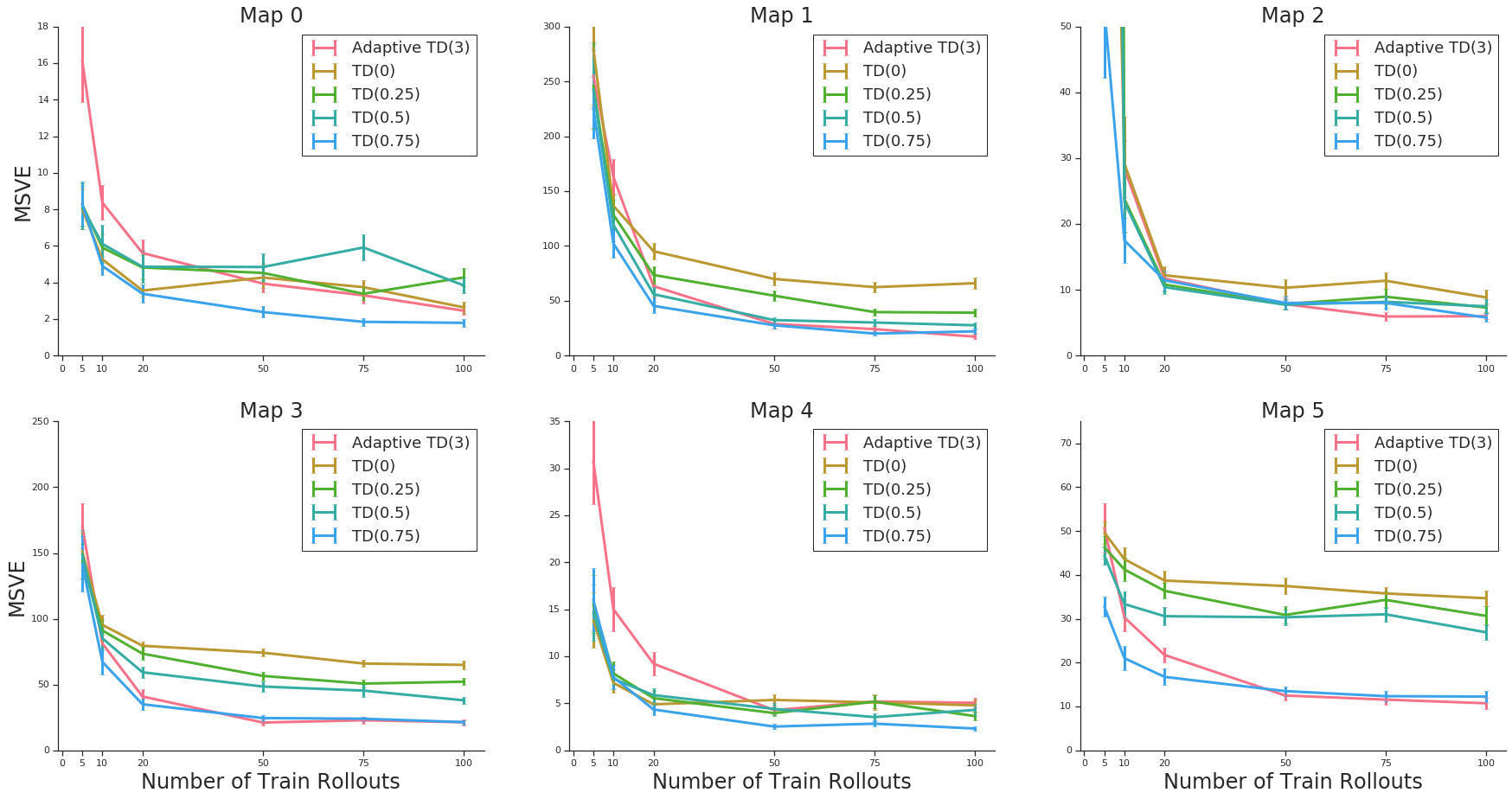}
\end{tabularx}
\caption{
\textbf{Labyrinth-2D.}
Unnormalized MSVE results for TD($\lambda$).
Confidence intervals over 20 seeds.
}
\label{fig:lab2_maps_lambdas_unnormalized}
\end{figure*}

In addition, we solved the same set of scenarios with a different function approximator: a simple piece-wise constant function on a 2D grid, where all the states within a cell are assigned the same value. Every map layout has size 400 x 300 and the grid layout chosen was of cells of size 19 x 19. This ensures that there was no easy coincidence with the grid layout and the position of the walls.

\begin{figure*}[h]
\setkeys{Gin}{width=\linewidth}
\begin{tabularx}{\columnwidth}{XX}
\includegraphics{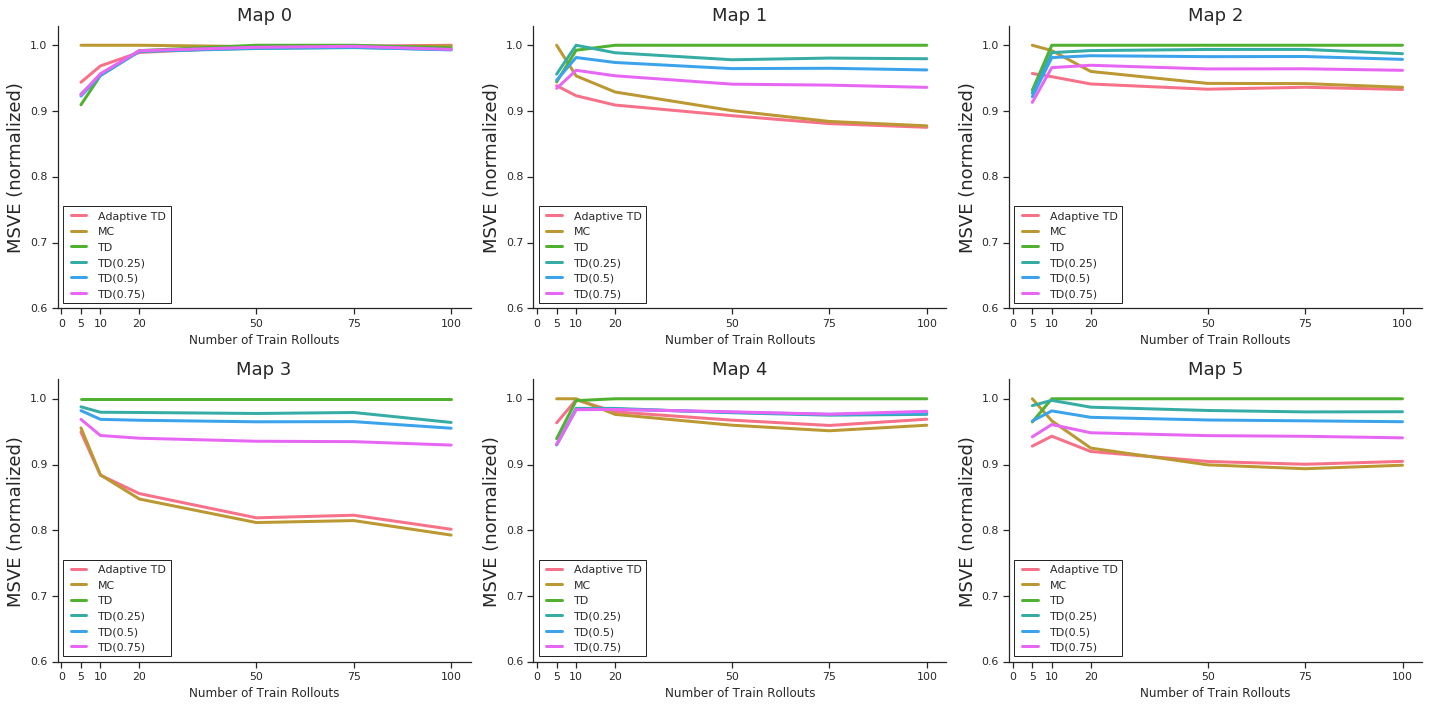}
\end{tabularx}
\caption{
\textbf{Labyrinth-2D.}
Algorithms use piece-wise constant approximation functions on a 2D grid.
For each number of train rollouts, we normalize the MSVE of each algorithm $A$ by $\MSVE(A) / \max_{A^\prime}\MSVE(A^\prime)$.
}
\label{fig:lab2_outcome_app_grid}
\end{figure*}

\begin{figure*}[h]
\setkeys{Gin}{width=\linewidth}
\begin{tabularx}{\columnwidth}{XX|XX}
\includegraphics[width=3cm,height=2.25cm]{labyrinth_2d/map_0.png} &
\includegraphics[width=3.5cm,height=2.25cm]{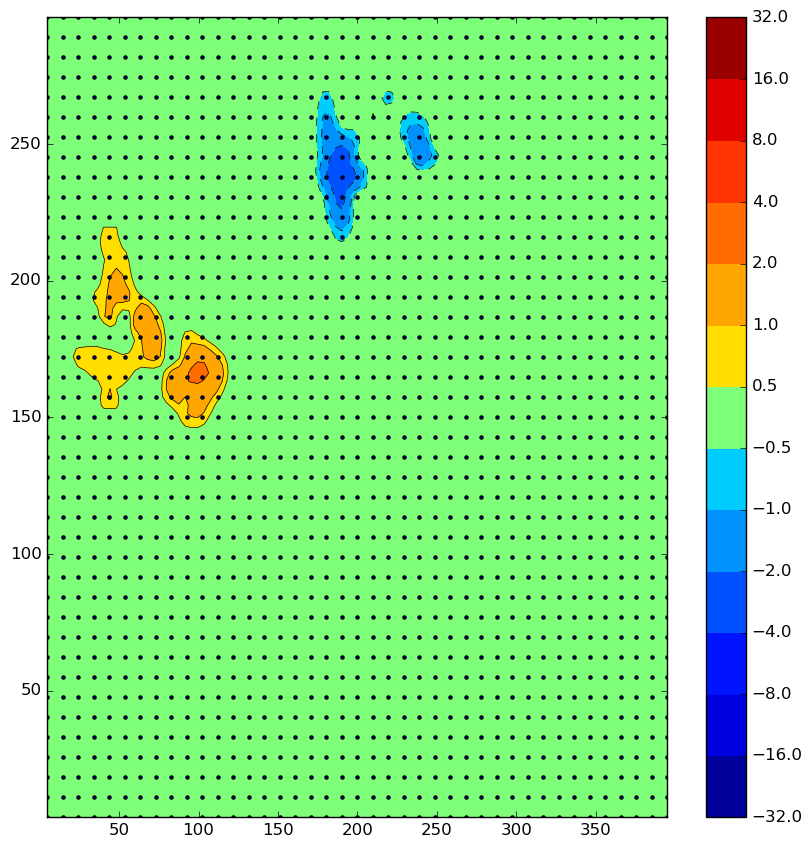} &
\includegraphics[width=3cm,height=2.25cm]{labyrinth_2d/map_3.png} &
\includegraphics[width=3.5cm,height=2.25cm]{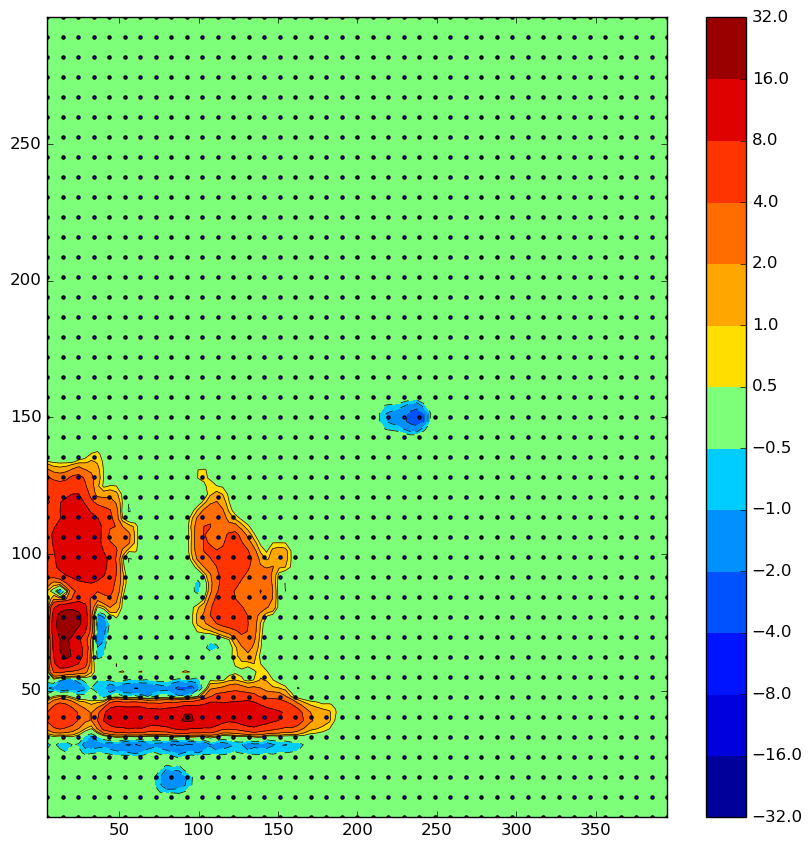} \\

\includegraphics[width=3cm,height=2.25cm]{labyrinth_2d/map_1.png} &
\includegraphics[width=3.5cm,height=2.25cm]{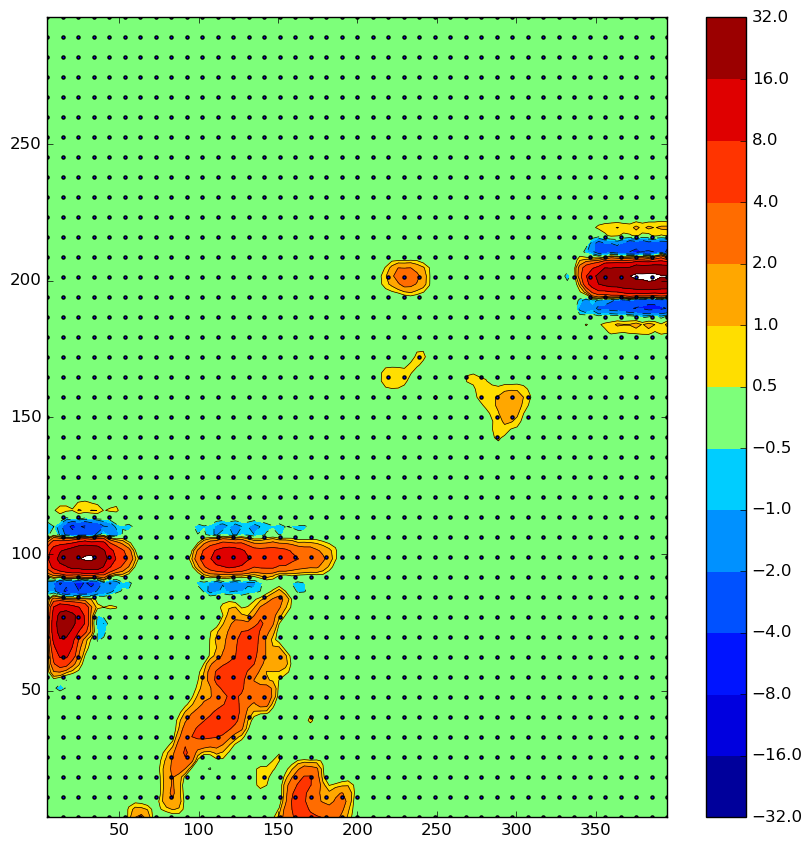} &
\includegraphics[width=3cm,height=2.25cm]{labyrinth_2d/map_4.png} &
\includegraphics[width=3.5cm,height=2.25cm]{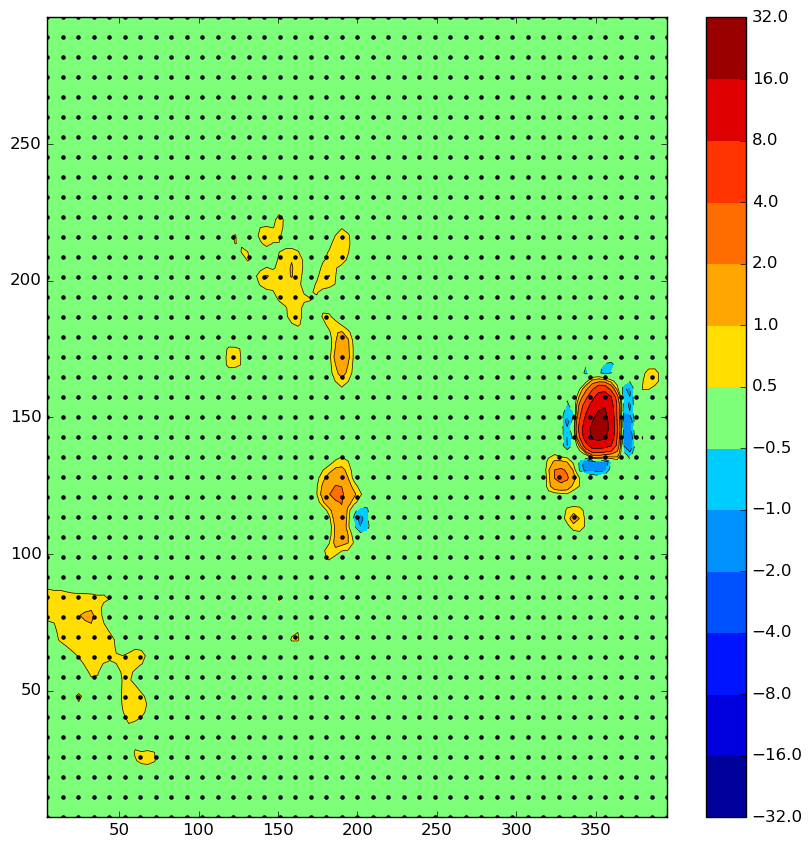} \\

\includegraphics[width=3cm,height=2.25cm]{labyrinth_2d/map_2.png} &
\includegraphics[width=3.5cm,height=2.25cm]{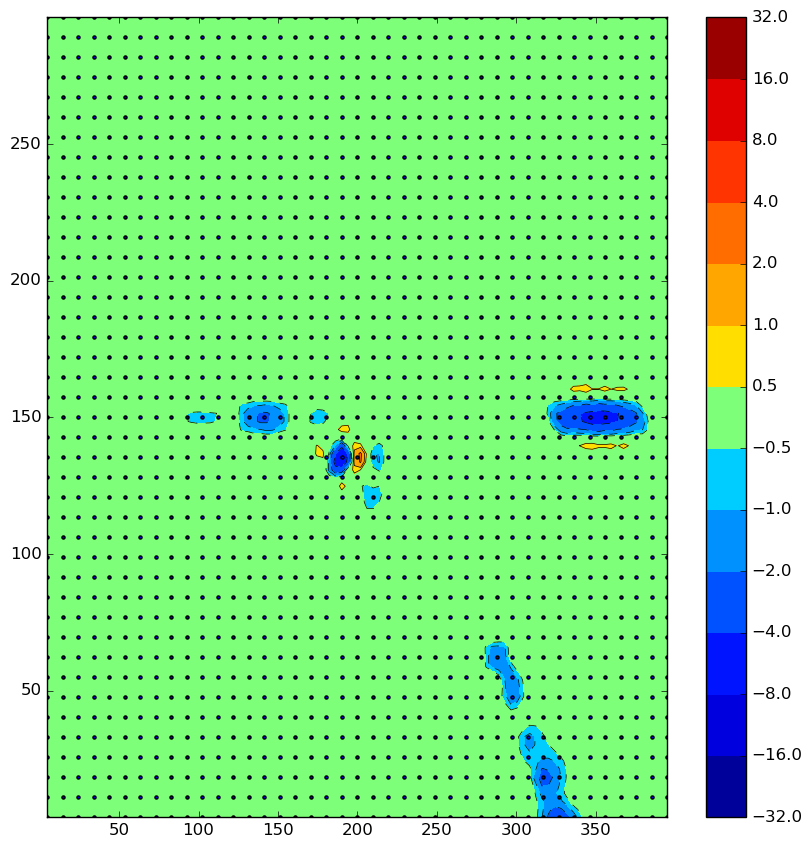} &
\includegraphics[width=3cm,height=2.25cm]{labyrinth_2d/map_5.png} &
\includegraphics[width=3.5cm,height=2.25cm]{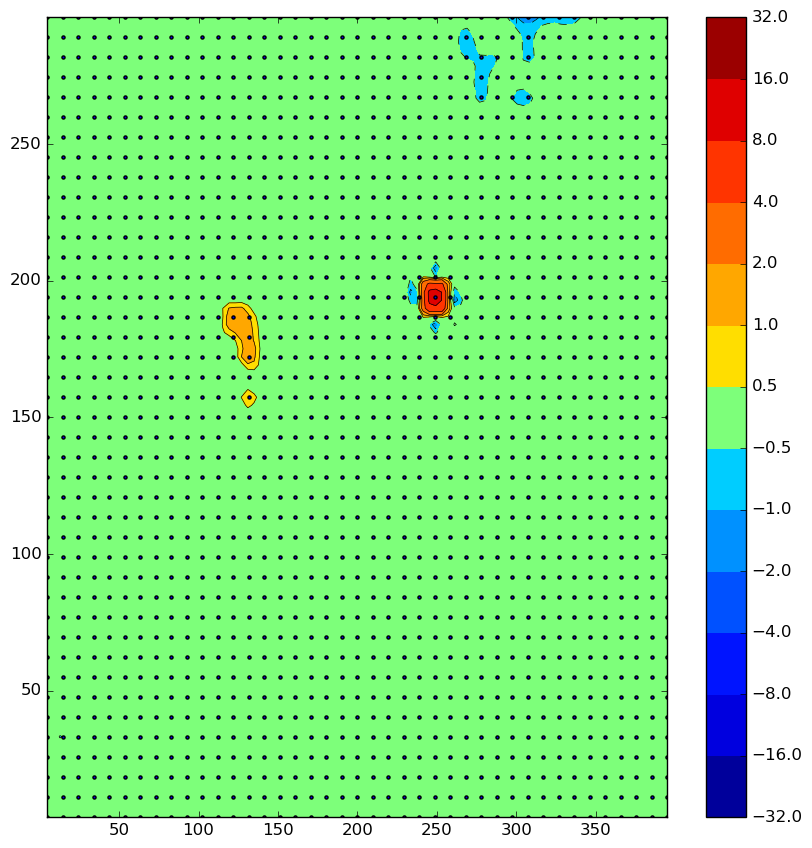} \\

\end{tabularx}
\caption{Visualization of the quality of the confidence intervals learned with the neural network ensembles. We take the ground truth value function and check for every state whether it falls inside the confidence interval (green color) or otherwise (red for over-estimation and blue for under-estimation). This shows that overall, the confidence intervals are reasonable.}
\label{fig:map2_ground_truth_conf_int_violations}
\end{figure*}

\clearpage
\section{Atari Environments}

For each of the 4 Atari games, we took a policy that was trained using DQN \cite{mnih2015human} on that game, and generated data by running it and sampling actions according to the softmax distribution of the output layer, with temperature = 0.02.
Each episode was limited to a maximum of 4000 steps. Training was done using varying amounts of episodes, from 5 to 200, as illustrated by the plots. For evaluation, in all cases, we used a disjoint set of 100 episodes generated using the same procedure. We use a discount factor of 0.99 in all experiments.

\begin{figure}[h]
\setkeys{Gin}{width=\linewidth}
\begin{tabularx}{\columnwidth}{XX}
\includegraphics{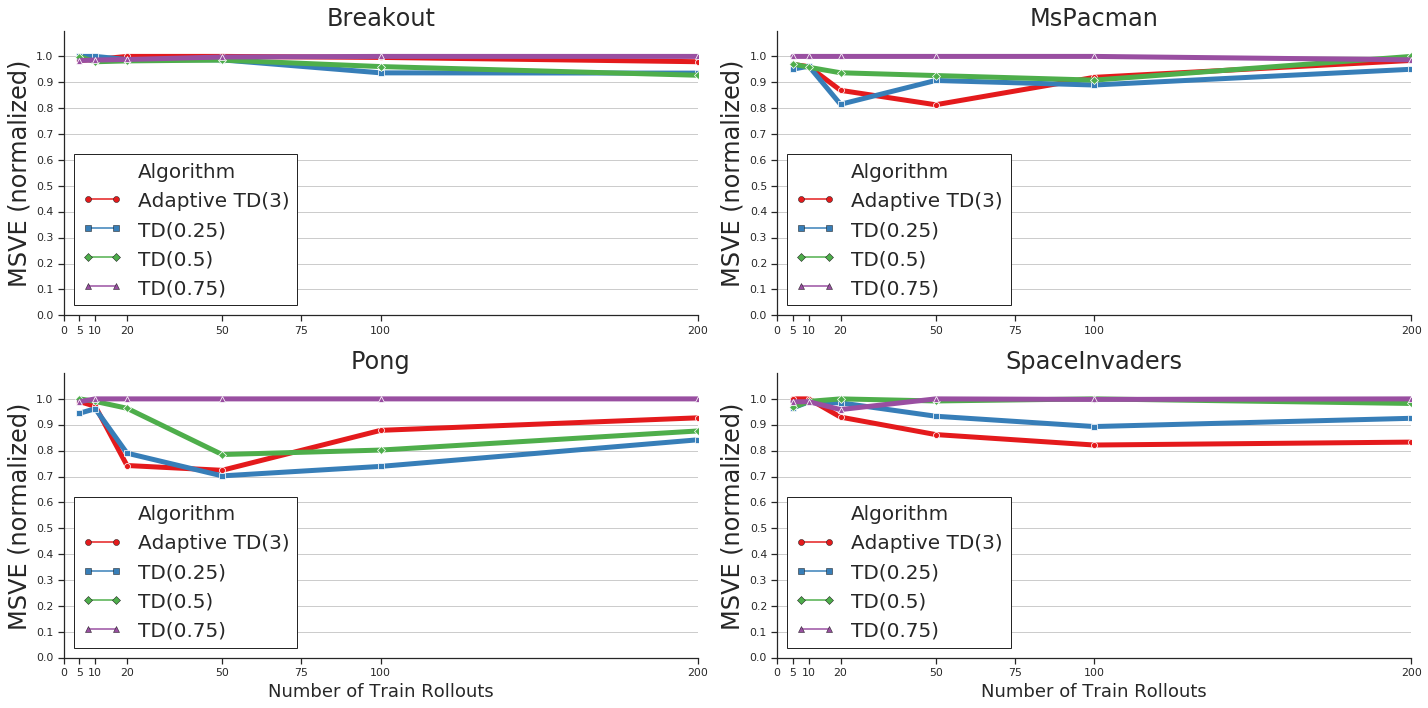}
\end{tabularx}
\caption{
\textbf{Atari.}
TD($\lambda$) results.
For each number of train rollouts, we normalize the MSVE of each algorithm $A$ by $\MSVE(A) / \max_{A^\prime}\MSVE(A^\prime)$.
}
\label{fig:atari_outcome_app}
\end{figure}

\clearpage

\begin{figure}
\setkeys{Gin}{width=\linewidth}
\begin{tabularx}{\columnwidth}{XX}
\includegraphics{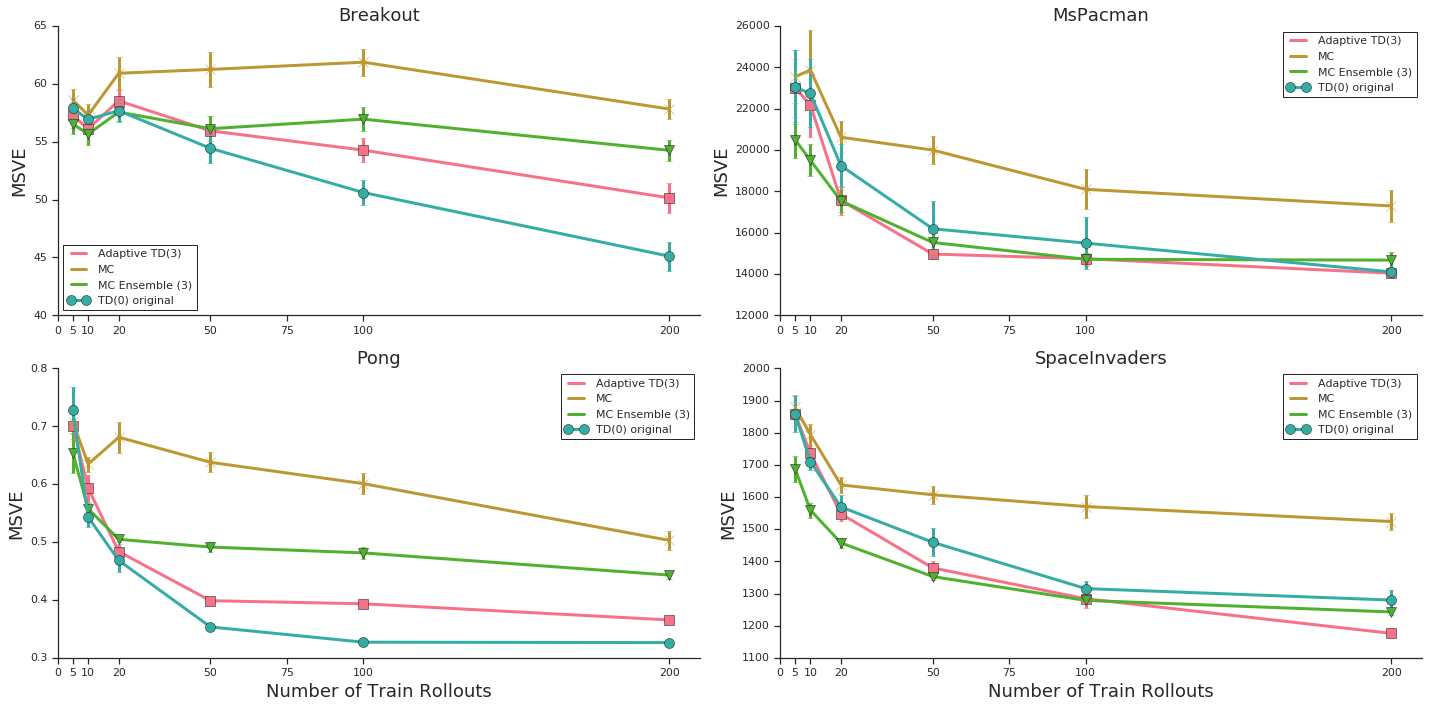}
\end{tabularx}
\caption{
\textbf{Atari.}
Unnormalized MSVE results for main baselines.
Confidence intervals over 20 seeds.
}
\label{fig:atari_outcome_app2}
\end{figure}

\begin{figure}
\setkeys{Gin}{width=\linewidth}
\begin{tabularx}{\columnwidth}{XX}
\includegraphics{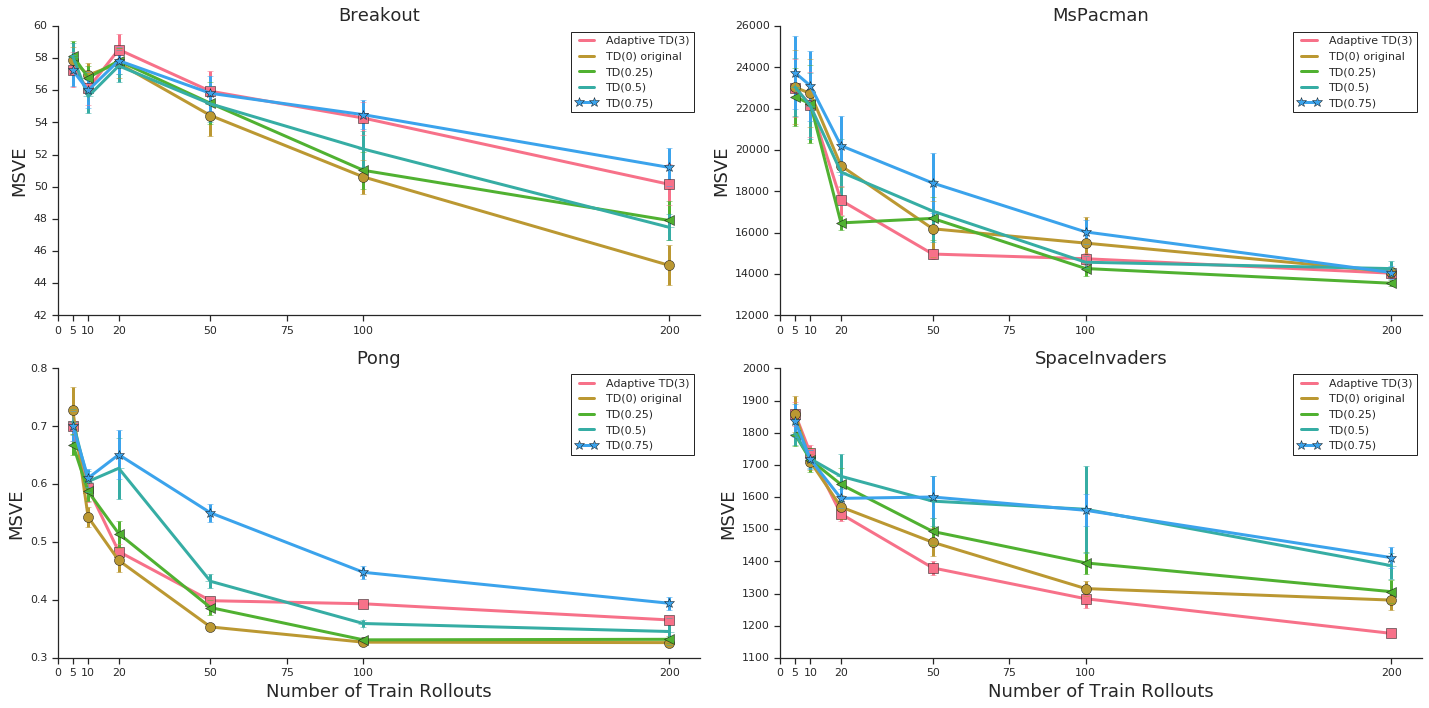}
\end{tabularx}
\caption{
\textbf{Atari.}
Unnormalized MSVE results for TD($\lambda$).
Confidence intervals over 20 seeds.
}
\label{fig:atari_outcome_app3}
\end{figure}

\end{document}